\documentclass[journal]{IEEEtran}

\usepackage{cite}
\usepackage{multirow}
\usepackage{soul}

\ifCLASSINFOpdf
  \usepackage[pdftex]{graphicx}
  \graphicspath{{images/}}
\else
\fi
\newcommand{\remindtext}[2]{{{#2}}}
\newcommand{\addedtext}[2]{{{#2}}}
\newcommand{\modifiedtext}[2]{{{#2}}}
\newcommand{\deletedtext}[2]{{}}

\begin{document}
\title{A Sensorized Multicurved Robot Finger with Data-\\driven Touch Sensing via Overlapping Light Signals}

\author{Pedro~Piacenza,
        Keith~Behrman,
        Benedikt~Schifferer,
        Ioannis~Kymissis,~\IEEEmembership{Member,~IEEE,}\\
        and~Matei Ciocarlie,~\IEEEmembership{Member,~IEEE}% <-this % stops a space
        \thanks{P. Piacenza and M. Ciocarlie are with the Dept. of Mechanical Engineering, Columbia University, New York, USA. K. Behrman and I. Kymissis are with the Dept. of Electrical Engineering, Columbia University. B. Schifferer is with the Dept. of Computer Science, Columbia University. Corresponding e-mail:pedro.piacenza@columbia.edu. Manuscript submitted August 2019.}}% <-this % stops a space
%\thanks{Manuscript submitted August 2019.}}


\markboth{IEEE Transactions on Mechatronics}%
{Piacenza \MakeLowercase{\textit{et al.}}}

\maketitle

\begin{abstract}
Despite significant advances in touch and force transduction, tactile
sensing is still far from ubiquitous in robotic manipulation. Existing
methods for building touch sensors have proven difficult to integrate
into robot fingers due to multiple challenges, including difficulty in
covering multicurved surfaces, high wire count, or packaging
constrains preventing their use in dexterous hands. In this paper, we
present a multicurved robotic finger with accurate touch localization
and normal force detection over complex, three-dimensional
surfaces. The key to our approach is the novel use of overlapping
signals from light emitters and receivers embedded in a transparent
waveguide layer that covers the functional areas of the finger. By
measuring light transport between every emitter and receiver, we show
that we can obtain a very rich signal set that changes in response to
deformation of the finger due to touch. We then show that purely
data-driven deep learning methods are able to extract useful
information from such data, such as contact location and applied
normal force, without the need for analytical models. The final result
is a fully integrated, sensorized robot finger, with a low wire count
and using easily accessible manufacturing methods, designed for easy
integration into dexterous manipulators.
\end{abstract}

%\begin{IEEEkeywords}
%IEEE, IEEEtran, journal, \LaTeX, paper, template.
%\end{IEEEkeywords}

% \fi
\IEEEpeerreviewmaketitle

\section{Introduction}

\IEEEPARstart{T}{actile} sensing modalities designed for robot hands have made great
strides over the past years. A number of comprehensive
reviews~\cite{dahiya2010,hammock2013,chortos2014,kappassov2015}
describe numerous tactile sensors, based on various transduction
methods (e.g., piezoresistance, piezocapacitance, piezoelectricity,
optics, ultrasonics, etc.). Still, these advances in sensing
modalities are only slowly translating to improved manipulation
abilities for robot hands. In particular, we posit that a gap that has
proven difficult to bridge has been that between stand-alone
\textit{tactile sensors} and fully integrated \textit{tactile
  fingers}.

To illustrate this difference, consider a robot hand operating in
cluttered, unstructured environments. Just like a stand-alone sensor,
a tactile finger should be able to collect and report rich data
characterizing touch. The information contained in the data will vary
depending on the application, but typical use cases require the
ability to infer touch location, characteristics of transmitted force,
or perhaps the motor actions to be applied by the robot in response to
the touch.

However, unlike stand-alone sensors, a tactile finger's performance is
also determined by the related problems of shape and coverage. When
operating in clutter, a finger equipped with discontinuous ``patches''
of tactile sensing, and ``unsensed'' areas of higher curvature in
between (such as edges or corners) has a high chance of making
contact in a blind spot. Furthermore, integration into a dexterous
manipulator (e.g., a multifingered hand with multiple tactile links on
each finger) places tight constraints on wiring and
packaging. Summarizing these goals, we are motivated to develop
robotic fingers exhibiting high tactile acuity over complex,
multicurved surfaces, with no ``blind'' edges or corners, in a
self-contained package with few wires, and ready for integration into
complete manipulators.

\begin{figure}[t!]
  \centering
  \setlength{\tabcolsep}{0mm}
  \begin{tabular}{c}
    \includegraphics[clip, trim=5cm 22cm 5cm 5cm,width=0.32\linewidth]{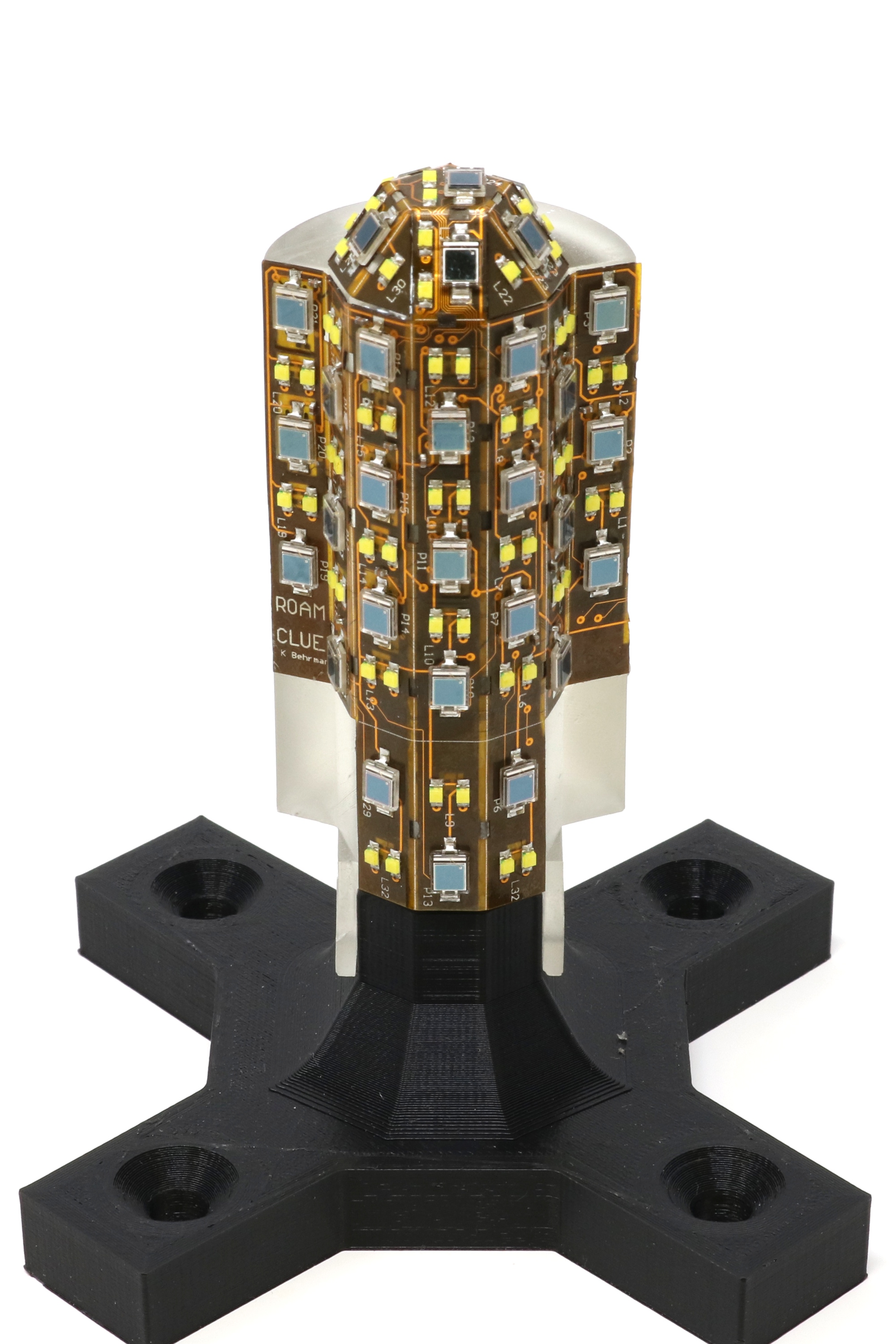}
    \includegraphics[clip, trim=5cm 22cm 5cm 5cm,width=0.32\linewidth]{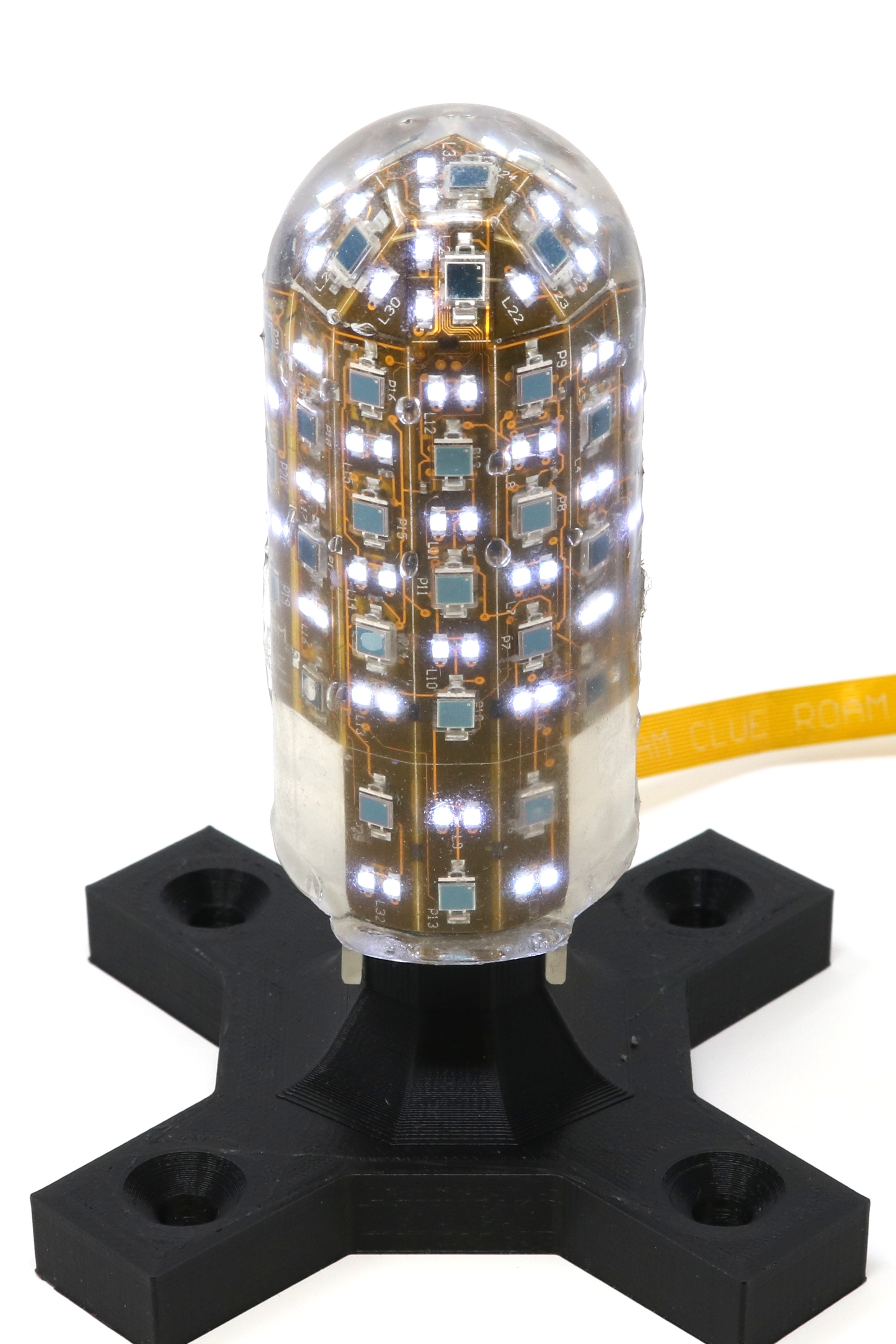}
    \includegraphics[clip, trim=5cm 22cm 5cm 5cm,width=0.32\linewidth]{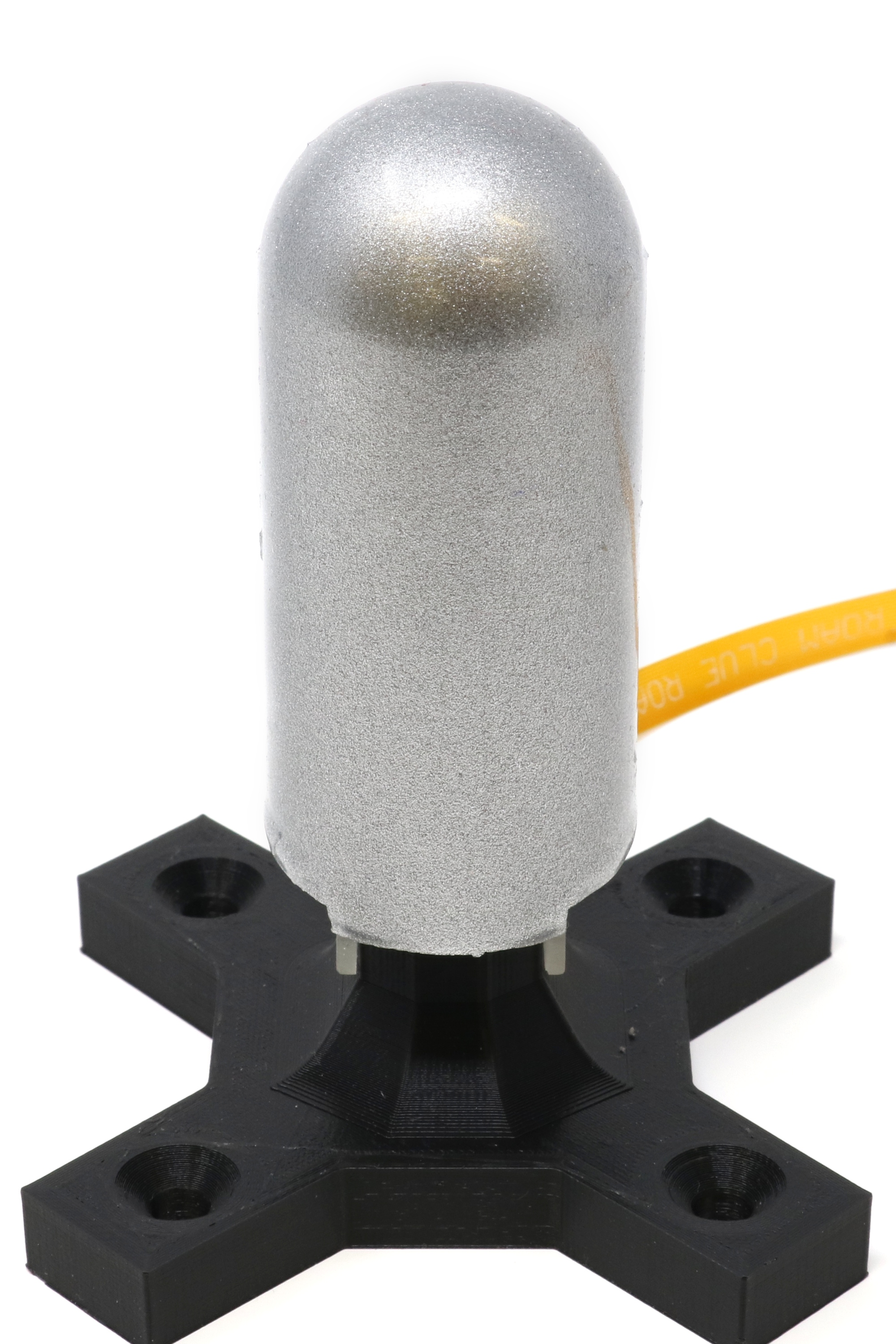}\\
    \includegraphics[clip, trim=1cm 0cm 1cm 0cm,width=0.5\linewidth]{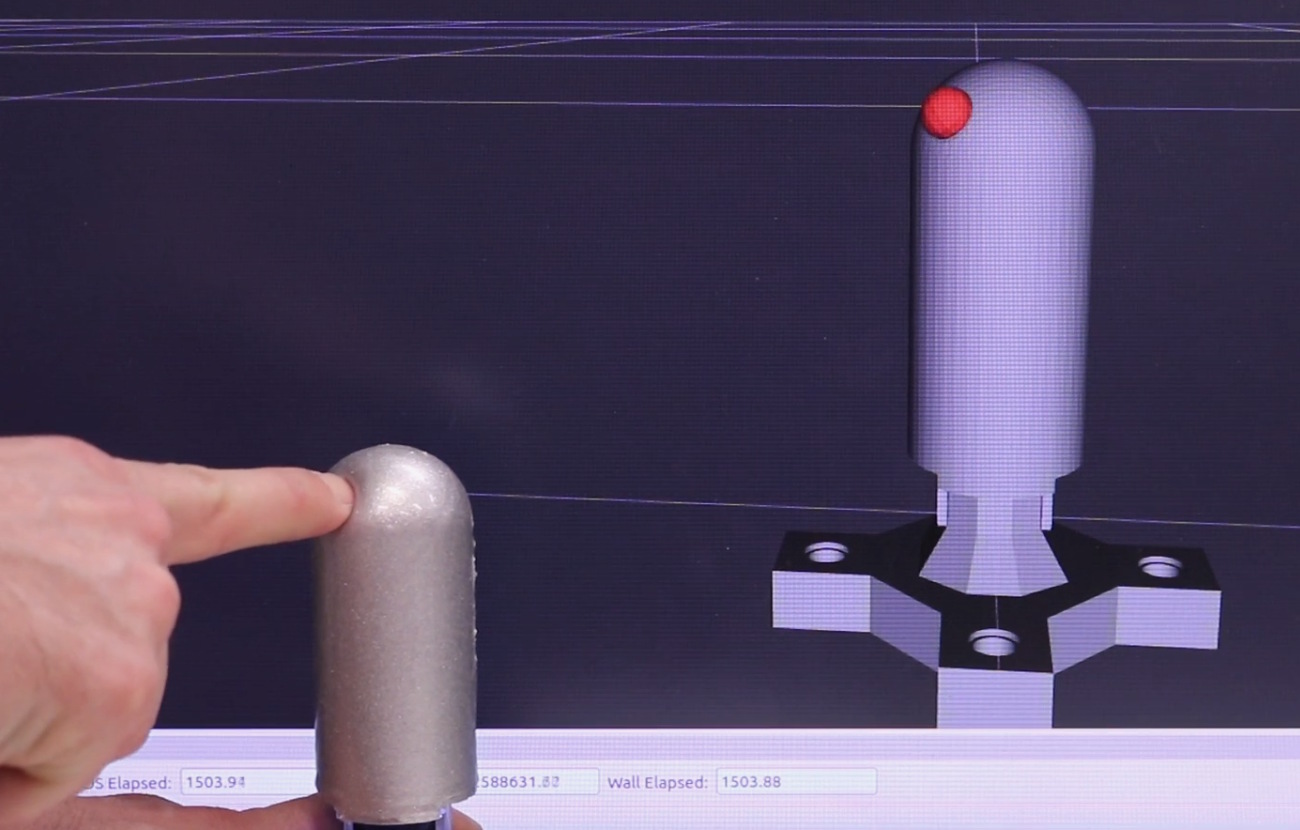}
    \includegraphics[clip, trim=1cm 0cm 1cm 0cm,width=0.5\linewidth]{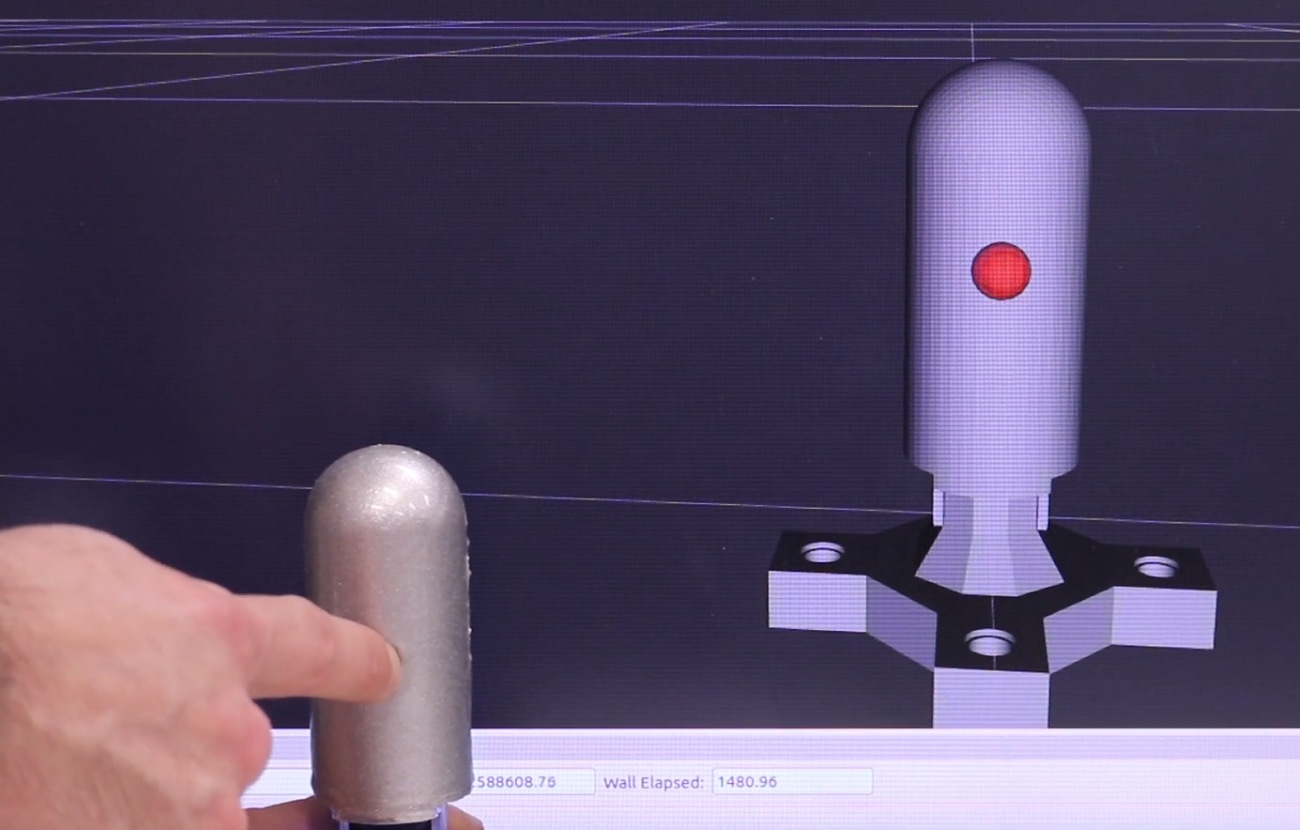}
  \end{tabular}
  \caption{A multicurved tactile finger. \textbf{Top:} finger through
    various stages of construction. We 3D-print a rigid skeleton, on
    which we attach a flexible circuit board with light emitters
    (LEDs) and receivers (photodiodes). We then mold a 7 mm thick
    transparent layer acting as a waveguide. Finally, we add a thin
    reflective outer layer. \textbf{Bottom:} finger performing touch
    localization and force detection. Location of red sphere shows
    predicted touch location, and sphere radius is proportional to
    predicted normal contact force.}
  \label{fig:finger}
\end{figure}

Traditionally, high accuracy contact localization has been achieved
using dense arrays of individual
taxels~\cite{kane2000,takao2006,suzuki1990}. Each taxel's response is
processed individually (e.g., isolated, linearized, de-noised) in an
attempt to prepare the data for upstream use, before it is offloaded
from the sensor. Recent work on e-skin has obtained
similar performance using soft materials resulting in flexible
sensors~\cite{ulmen2010,mittendorfer2011,chortos2014,buscher2015,boutry2018,shimojo2004,kim2008,byun2018,osborn2018,wu2018,drimus2014}. However,
reading high quality signals from individual taxels comes at a cost of
increased manufacturing complexity, as taxels must be isolated from
each other, high resolution arrays result in increased wiring, and
deploying these sensors on non-planar and non-developable surfaces
increases manufacturing complexity. In contrast to the use of
carefully calibrated and isolated signals, recent years have seen
increased adoption in robotics of machine learning methods that are
well suited at quickly processing large amounts of ``raw'' data, with
little to no pre-processing. We believe that this perspective can be
used to relax some of these previously held assumptions (such as the
``one taxel, one signal'' approach), and enable new approaches to
tactile sensing.

%\subsection{Overlapping Optical Signals.}

We base our work on the key concept of overlapping optical signals. To
illustrate this concept, consider one light emitter (in our case, an
LED) and one light receiver (a photodiode) embedded in a transparent
medium acting as a waveguide. Any deformation of this medium (e.g., due
to contact with an external indenter) will change the amount of light
traveling from the LED to the diode, providing a measurable signal
that is related to contact with the indenter.

Taking this concept further, consider multiple light emitters and
receivers embedded in the same medium. Any external touch which
deforms the medium will lead to changes in signal between multiple
LED/diode pairs. In fact, assuming an equal distribution of $n$ LEDs
and $n$ diodes, the total number of signals we can measure is $n^2$,
or the amount of light transmitted between each LED and each
photodiode. Because each LED/diode pair has its own spatial receptive
field within the active sensing area, and these receptive fields
overlap, we call this general method \textit{spatially overlapping
  signals}. The total number of signals we can collect is quadratic in
the number of individual sensing units (LEDs and diodes), giving us
extremely rich data with relatively few wires.

We note that the use of spatially overlapping signals is not
necessarily restricted to using optics as an underlying transduction
method. Other methods can be used, such as piezoresistance or
electrical impedance. We chose light transport in our work due to
attractive properties such as simple and low-cost manufacturing (the
medium consists simply of a transparent polymer, as opposed to
piezoresistive materials), low hysteresis, and fast switching and
sampling. 

The finger we demonstrate in this work (Fig.~\ref{fig:finger})
comprises a hemispherical tip attached to a cylindrical body. The
sensorized areas of the finger include the complete hemispherical tip,
and half the circumference (180$^{\circ}$) of the cylindrical
body. The finger contains 32 LEDs and 30 photodiodes, giving us a
total of 960 signals. As the finger makes contact with an external
surface, many of these signals change, with the sign and magnitude of
the change for each signal depending on the deformation of the
transparent medium, as well as the relative position of the respective
LED and diode.

%\subsection{Data-driven Tactile Sensing}

How can we make use of this rich dataset? Building an analytical
model of how each signal is affected by contact characteristics is a
daunting task; furthermore, any such model would depend on knowing the
exact locations of the terminals in the sensor, thus requiring very
precise manufacturing. In contrast, we propose a purely data-driven
approach, where the mapping from our signals to the quantities of
interest is learned directly from data. This approach is enabled by
recent advances in machine learning allowing us to train regressors
and classifiers on high-dimensional feature spaces, as is the case for
our tactile finger.

A data-driven approach needs ground truth for training. We thus
collect labeled ground truth data by indenting our fingers in
controlled conditions, using an indenter mounted on a robot arm and
equipped with a load cell. For any indentation, we record the exact
location of contact (based on the robot arm encoders) and the normal
force (reported by the load cell). For multitouch, we use a manual
procedure to record the identity of discretized finger cells being
touched. We use this dataset to train models for predicting contact
location(s) and normal force based on the optical signals recorded by
the finger. We use deep neural networks for both regression (for
location and force) and classification (multitouch detection).

We summarize the contributions of the this paper as follows: To the
best of our knowledge, we introduce the first multicurved robotic
finger that can localize touch with sub-millimeter accuracy and also
accurately determine normal contact force over non-developable
three-dimensional surfaces (such as a cylindrical body with a
hemispherical tip), in a fully integrated, finger-shaped package. To
achieve this, we show that purely data-driven methods can extract
useful information from overlapping optical signals, enabling fast
operation, easily accessible manufacturing methods, and a low wire
count for the integrated finger.

\section{Related Work}

Optics-based tactile sensing has a long history of integration in
robotic fingers and hands~\cite{begej1988,schneider2009,li2019}. Of
particular interest is the use of CCD or CMOS sensors recording light
patterns through a robotic tip. This includes the
GelSight~\cite{johnson2009}, TacTip~\cite{lepora2015_superresolution}
and GelSlim~\cite{ma2019} sensors, which can retrieve minute details
of surface texture, and also achieve super-resolution and
hyperacuity. However, the imaging array must image the entire touch
area, leading to bulky assemblies or partially sensorized fingers. In
our work, the sensing terminals are fully distributed, allowing for
coverage of large areas and curved geometry. Waveguides are also used
as force transducers~\cite{polygerinos2010,konst2017} with very high
sensitivity, but without contact localization.

\modifiedtext{1-2}{Previous work also overlapped multiple optical
  signals in the same waveguide~\cite{levi2013} in order to enable a
  reconstruction of the applied pressure map, but no quantitative
  results on reconstruction or localization accuracy were
  presented. Other studies used a one-to-many~\cite{yun2014} or
  one-to-one~\cite{quigley2014} paradigm for signals between light emitters and
  receivers, as opposed to the many-to-many paradigm we use here in
  order to increase localization accuracy.}  Reflection and refraction
were also used to build an IR touch sensor that doubles as a proximity
~\cite{patel2017} or proprioceptive sensor~\cite{meerbeck2018}, but
without localizing contact.

Extracting contact information from overlapping signals has been done
in the context of methods inspired by Electric Impedance Tomography
(EIT)~\cite{nagakubo2007,kato2007,tawil2011_improved,lee2017}; for a
comprehensive survey on the use of EIT for robotic skin see the work
of Silvera et al.~\cite{silvera2015}. EIT offers stretchable,
continuous tactile sensing with the ability to discriminate multiple
points of contact. An intrinsic advantage of EIT is that it can
produce full contact maps for multi-touch situations, an ability which
we have not yet investigated with our method. However, EIT methods
require an analytical model for the internal conductivity to construct
an image showing the areas where strain is applied. In contrast, our
approach is completely data-driven and does not require knowledge
of a forward model of the sensor, which allows us to embed terminals
along complex multicurved boundaries of three-dimensional sensing
areas.

Other sensors also use a small number of underlying transducers to
recover richer information about the contact, using super-resolution
or related
methods~\cite{heever2009,lepora2015_superresolution,lepora2015_tactile,muscari2013}. In
comparison, our data-driven approach allows arbitrary placement of
transducers inside the finger, and thus coverage of complex
multicurved surfaces. Random location of transducing terminals in a
soft finger has been used for texture
discrimination~\cite{hosoda2006}, but without the ability to localize
contact or measure force. The BioTac sensor~\cite{wettels2008}
pioneered an overlapping-signals approach for tactile sensing by
measuring impedance changes in a fluid, in addition to other
multimodal data (fluid temperature and pressure changes due to
contact). However, impedance signals were measured between a single
ground electrode and multiple working electrodes, producing a number
of signals that was linear in the number of electrodes, as opposed to
our approach of measuring an optical signal between any emitter and
any receiver, and thus obtaining a number of signals quadratic in the
number of terminals.

In the absence of learning-based or super-resolution techniques, high
spatial acuity has been achieved through high-resolution taxel
arrays~\cite{kane2000,takao2006,suzuki1990}, which cannot cover
curved surfaces. The highly active field of e-skin
research~\cite{ulmen2010,mittendorfer2011,
  chortos2014,buscher2015,boutry2018,shimojo2004,kim2008,byun2018,osborn2018,wu2018,drimus2014}
shows promise to overcome such problems through flexible sensors that
include features like high density of sensing elements, multimodal
sensing, and even actuation~\cite{booth2018}. Flexible tactile skin
has also been used to sensorize gloves for collecting tactile data on
human manipulation~\cite{buscher2015}, and taxelized skins can
inherently distinguish multitouch conditions. However, manufacturing
complexity, along with system-level issues such as wiring, addressing,
signal processing of multiple sensor elements or off-board
amplification electronics remain important roadblocks on the way to
using e-skin for fully integrated sensorized robot fingers exhibiting
sub-millimeter localization accuracy as well as complete coverage of
complex non-developable geometry, as the one we present here.

Machine learning methods are seeing increased used for manipulation
based on tactile data. Examples include grasp adaptation or object
identification through tactile
sensing~\cite{hogan2018,drimus2014,li2014,allen2013},
slip and rotation detection~\cite{meier2016}, learning to discriminate
between different types of geometric features~\cite{wong2014} using
the BioTac sensor~\cite{wettels2008}, learning the mapping between
tactile signal variability and grasp stability~\cite{wan2016} with the
MEMS-based Takktile sensors~\cite{tenzer2014}, using recurrent neural
networks for proprioception on soft manipulators~\cite{thuruthel2019},
and classifying different touch gestures on a novel EIT-based
skin~\cite{tawil2011_interpretation}. Here, we apply data-driven
methods to learn a model of the sensor itself and believe that
developing the sensor simultaneously with the learning techniques that
make use of the data can bring us closer to achieving complete tactile
systems.

\section{Tactile Finger Design and Manufacturing}

Our tactile robotic finger is constructed using LEDs and photodiodes
embedded in a clear elastomer which acts as a waveguide. Each possible
combination of an LED with a photodiode is considered a sensing
pair. Our method relies on the fact that, when the elastomer is
deformed under contact, the surface normal is perturbed and light is
deflected, producing a measurable change in the output of one or more
photodiode(s). Each LED/photodiode pair has a signal associated to it,
represented by the amount of light received by the respective
photodiode from the respective LED, and a receptive field, represented
by the area on the elastomer that can be disturbed to produce change
on this signal. Because we have many LEDs and photodiodes, and light
can travel a considerable distance inside the elastomer, the receptive
fields of different pairs overlap significantly. A single contact can
stimulate many different sensing pairs which is why we refer to this
methodology as spatially overlapping signals. Note that this method
gets rid of the traditional notion of a taxel, as we produce signals
with a combination of an emitter and receiver without enforcing a
particular location for either one.

\subsection{Finger geometry and surface parameterization}
We start with a 3D printed skeleton (Formlabs Form2, clear resin)
designed to be the distal link in a robot hand. This skeleton provides
a base on which we will mount our individual sensing terminals (LEDs
and photodiodes). The geometry of our finger is informed by the
requirements of the task as well as intuition. We chose the outer
geometry (including the transparent layer) to be appropriate for a
finger operating in clutter, maximizing the sensorized area while
avoiding sharp edges and corners, which are not conducive to creating
stable contacts. The outer shape thus consists of a 36 mm diameter
hemisphere, mounted on a cylindrical base of the same diameter and 72
mm height.

Our goal for the functional area of the finger (the area where touch
can be sensed) was to avoid any blind spots in areas likely to make
contact. We thus selected as an active area the complete tip
hemisphere, as well as 180$^{\circ}$ around the circumference of the
cylindrical body. The only ``unsensed'' area of the finger is the
back (half the circumference) of the cylindrical part. The transparent
waveguide layer extends throughout the sensorized area of the
finger. Based on previous experiments on planar
prototypes~\cite{piacenza2017}, we selected 7 mm as the thickness of
the transparent layer.

\begin{figure}[t]
  \centering
  \begin{tabular}{c}
    \includegraphics[height=50mm]{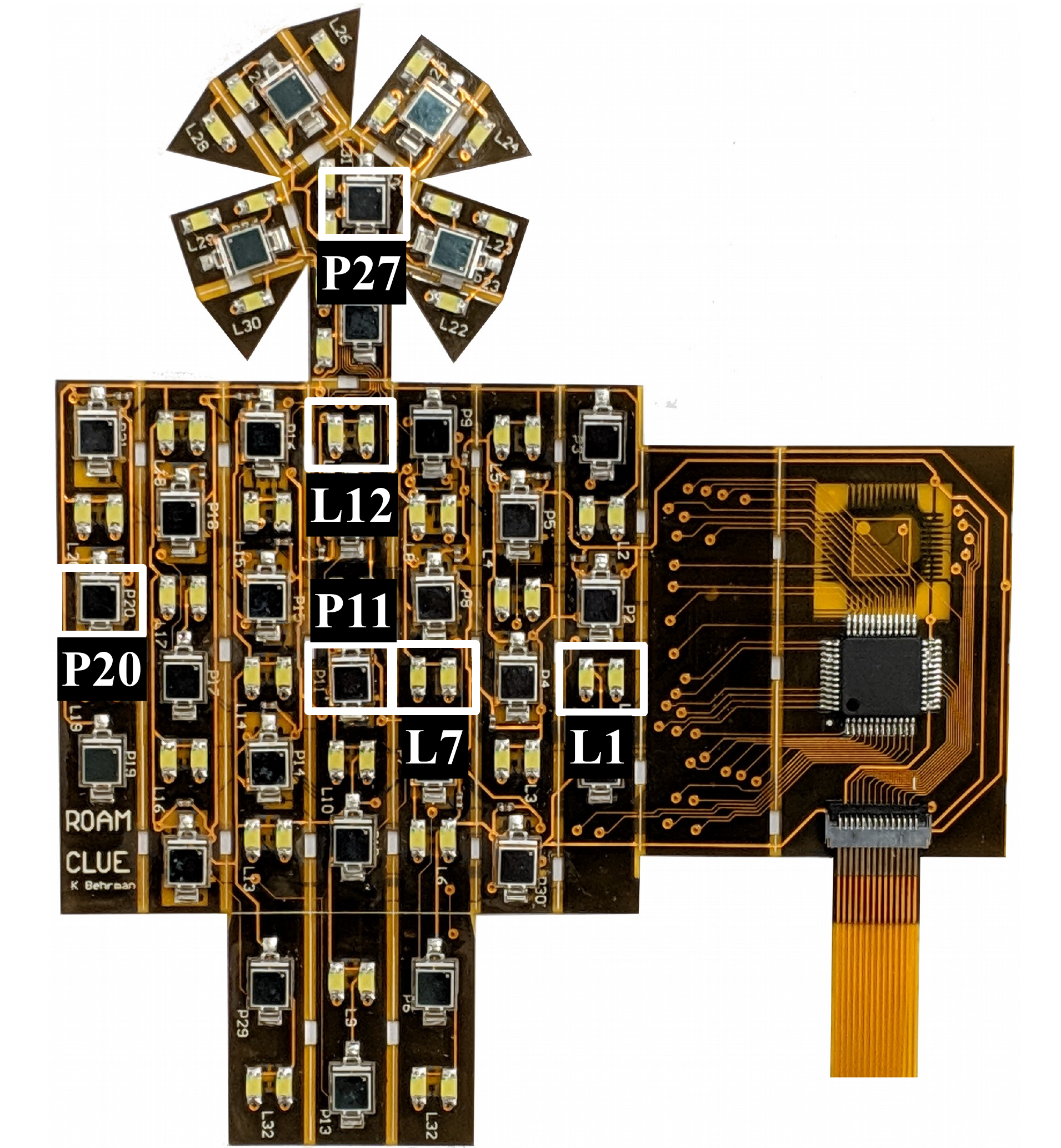}
    \includegraphics[clip, trim=12cm 2.3cm 8.5cm 1cm, height=50mm]{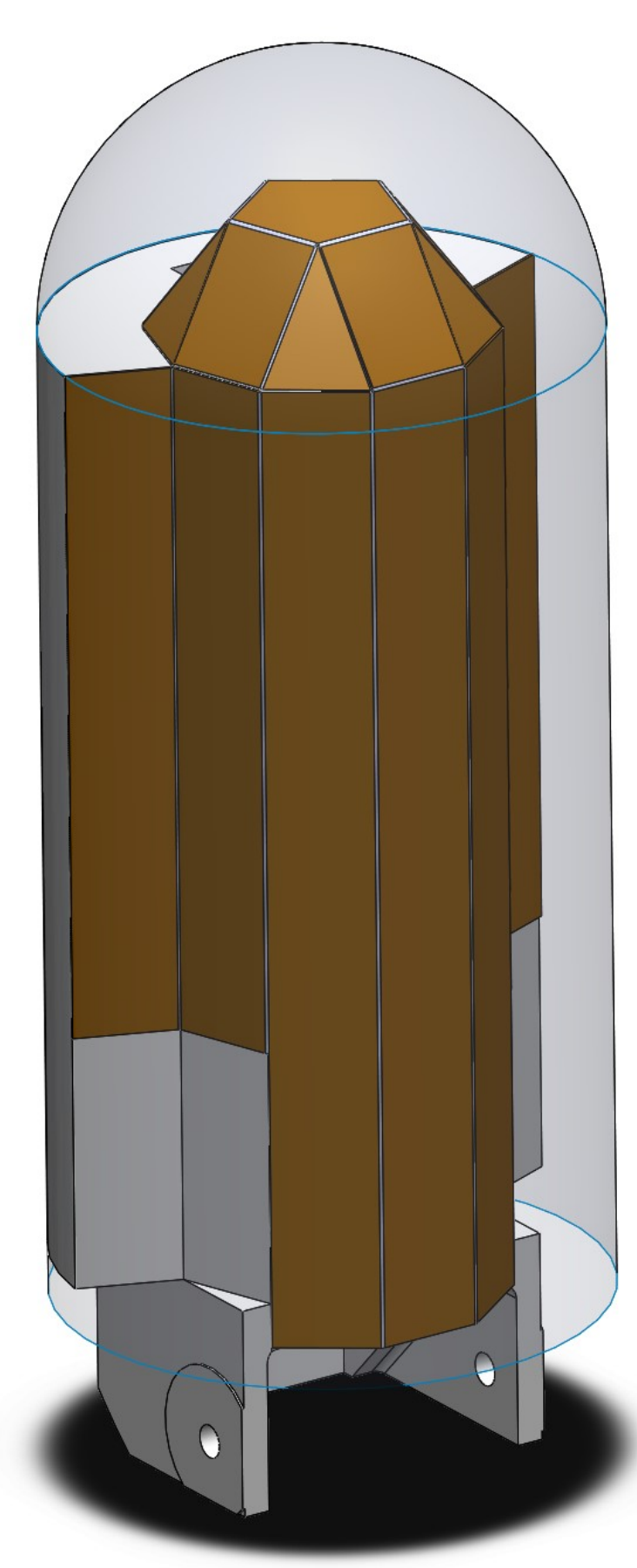}\\[2mm]
    \includegraphics[scale=0.1, clip, trim=7cm 2cm 11cm 2cm,height=40mm]{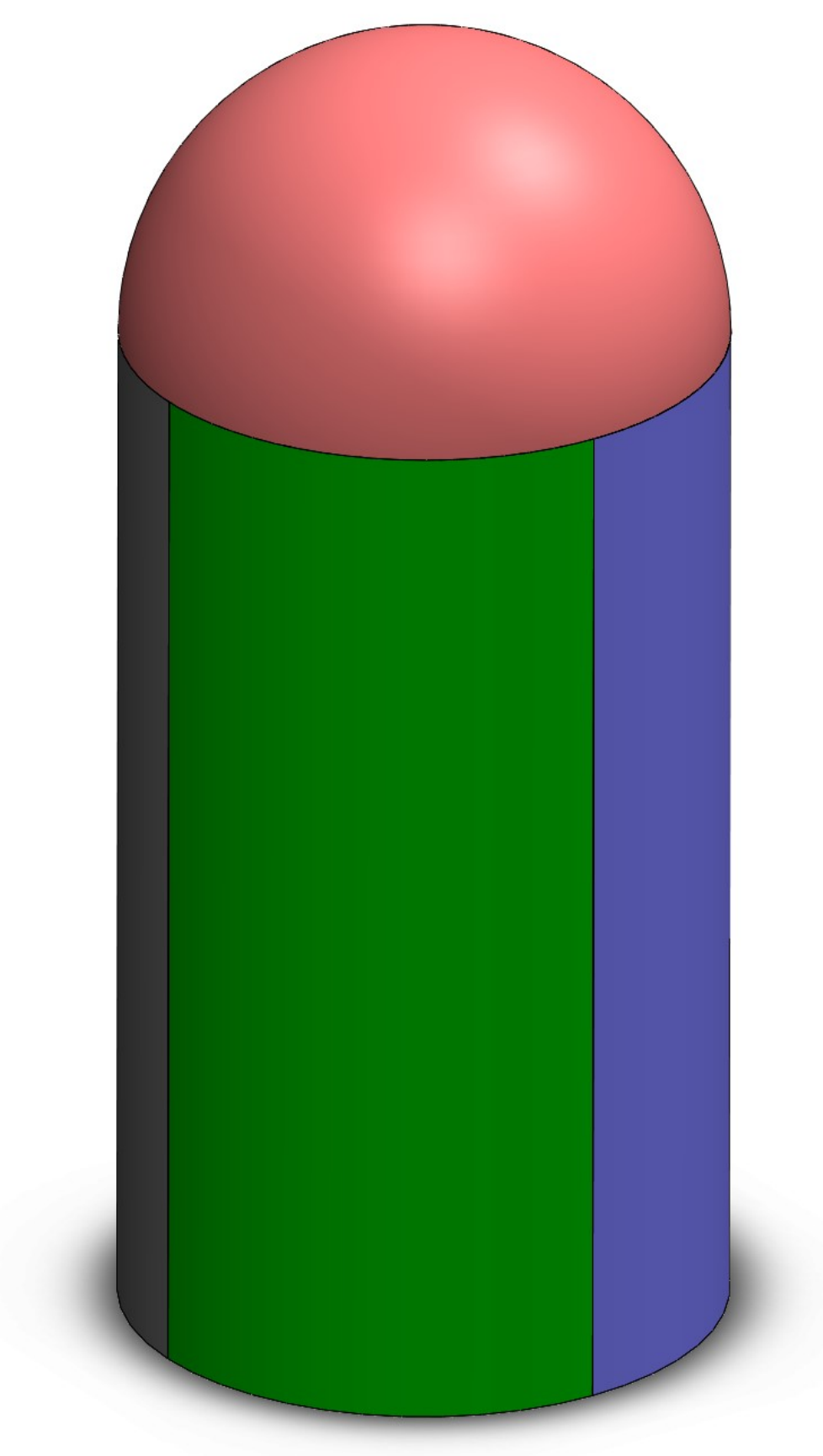}
    \raisebox{20mm}{\Large$\longrightarrow$}~~
    \includegraphics[clip, trim=0.3cm 0.5cm 0.3cm 0.3cm,height=40mm]{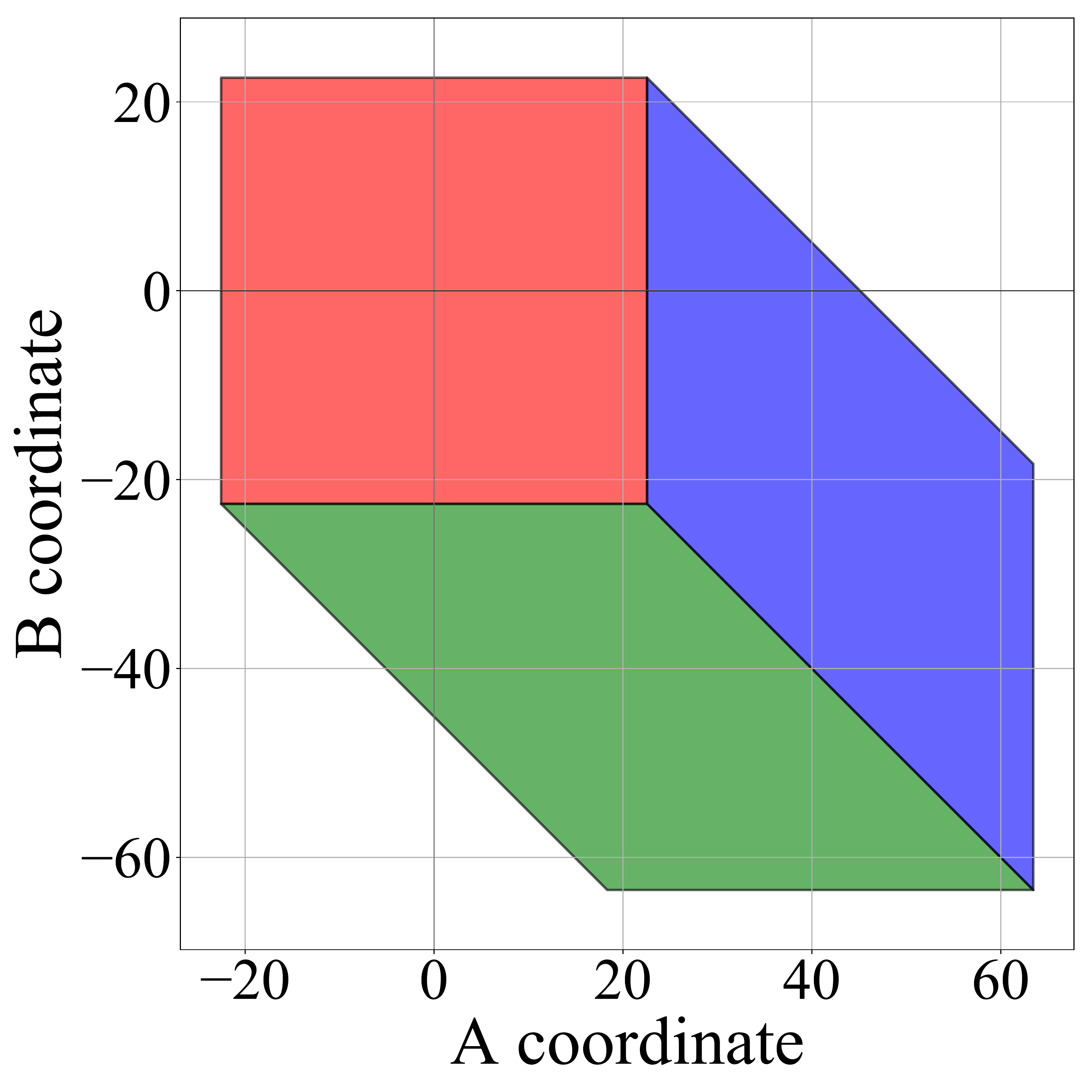}
  \end{tabular}
  \caption{Finger geometry. \textbf{Top:} flexboard with LED and
    photodiodes, as well as CAD showing flexboard wrapping on rigid
    finger skeleton and covered by 7 mm transparent layer. Three LEDs
    (L1, L7 and L12) and three diodes (P11, P20 and P27) are
    identified, as they will be referenced in later
    figures. \textbf{Bottom:} Finger surface parameterization into
    two-dimensional $(A,B)$ space. Finger is divided into three
    color-coded areas; each area maps to its corresponding color in
    $(A,B)$ space. Parameterization is continuous throughout the
    functional area of the finger, has no singularities, and attempts
    to preserve local surface area.}
\label{fig:geometry}
\end{figure}

The rigid skeleton acts as a foundation for the transparent layer, as
well as a support for the LEDs and photodiodes. The electronic
components are mounted on the flexible circuit board (flexboard) shown
in Fig.~\ref{fig:geometry}. Since most components require locally flat
mounting areas, the flexboard alternates flat strips (enforced with
stiffeners) with flexible ``creases''. The rigid skeleton is designed
with flat facets to accommodate this flexboard geometry. These flat
faces follow the smooth curvature of the finger surface above, and
thus the electronic components radiate and receive most light in the
same direction as the surface normal. \modifiedtext{3-5}{The left-most
  and right-most faces are the exception to this rule}, as they are
oriented such that LEDs will radiate light parallel to the surface
instead of normal to it. Light emitted in this way can travel all the
way to the other side of the finger as it bounces through the
reflective coating, increasing spatial coverage for a subset of
signals.

Since the surface of our finger is a (multicurved) two-dimensional
manifold (embedded in three dimensions), we parameterize it using two
dimensionless variables $(A,B)$. We use the surface parameterization
introduced by Ro{\c{s}}ca\cite{rocsca2010} for spheres, which we
extend to cover the cylindrical component of our finger (see
Appendix). The resulting parameterization, illustrated in
Fig.~\ref{fig:geometry}, avoids singularities and aims to preserve
distances uniformly across the sensorized surface of our finger.

\subsection{Manufacturing}

We use a cyanoacrylate adhesive to bond the flexboard to the skeleton
and then connect a 380 mm flat flexible cable (FFC) to the backside of
the finger. Once the flexboard is glued, we treat all the surfaces
with a silicone primer (MG Chemicals SS4120) to promote bonding of the
transparent waveguide layer to the skeleton and sensor board.

We then cast the transparent layer, for which we use
Polydimethylsiloxane (PDMS). \modifiedtext{2-4}{The stiffness of the
  elastomer can be adjusted by changing the ratio of curing agent to
  PDMS. We prepare the main layer of elastomer with a ratio of 1:30,
  which we empirically observed to produce a stiffness of
  approximately 2.8 N/mm.} We consider it desirable for this first
layer to be soft because our signals are proportional to surface
deformation; thus, the softer the material, the less force is required
to achieve a deformation that our electronics can detect. After mixing
the PDMS with the curing agent, we degas and pour the solution into
the mold. We cure the PDMS at 80 C for 6 hours. After this process,
we remove the finger from the mold to obtain a fully clear finger as
shown in Fig.~\ref{fig:finger}.

\remindtext{3-3}{To complete the finger, we add a thin outer layer
  with a dual purpose: this layer reflects light back into the transparent
  elastomer that otherwise might refract out, and blocks ambient
  light.} Following the example of the GelSight
sensor~\cite{johnson2009}, we use Bronzing Powder \#242 from Douglas
and Sturgess. After demolding the finger we use a makeup brush to
apply the reflective powder onto the PDMS surface, then add an
additional thin PDMS layer. This additional layer has a 1:10 ratio of
curing agent, as well as 1\% by weight reflective powder mixed in. We
add this layer via dipping the finger in a vat, followed by curing at
80 C for 40 minutes. The process of adding the reflective powder onto
the PDMS, followed by an extra PDMS layer to protect it is repeated
two more times.

\subsection{Sensor board and electronics}

The flexboard contains 57 individual LEDs (Bivar SM0805UWC), 30
photodiodes (Osram BPW 34 S E9601), 15 operational amplifiers
(AD8616), two 32 channel multiplexers (ADG732) and one FFC connector
with 14 positions (Hirose FH34SRJ-14S). Many of the Bivar LEDs are
grouped in pairs, comprising 2 units placed 2 mm apart and wired
together. Such a pair is the logical and functional equivalent of a
single LED with a larger surface area. We refer to such pairs as
simply ``one LED'' for the rest of the paper. Using this convention,
our board comprises a total of 32 ``logical''
LEDs. \addedtext{1-1}{Since we only turn on a single LED pair at a
  time (see below), and with each LED drawing 20 mA, our board
  uses a total of 200 mW power during operation.}

Similarly, we selected our photodiode for its large active sensing
area of 7 mm$^{2}$. We aimed for large surface areas in both our light
emitters and receivers in order to obtain a smooth response as light
paths are altered by the presence of an indenter (as opposed to a
binary-like signal that occurs when a small area emitter or receiver
is blocked or revealed). Each photodiode signal is amplified using a
transimpedance amplifier with a feedback resistor of 249 kOhms.

\modifiedtext{3-4}{In order to measure the light transmitted between
  each LED and each diode individually, we use one multiplexer to
  select which LED to drive and the other multiplexer to select which
  photodiode to read. We turn on one LED at a time, and read the
  signals from each photodiode in sequence.} We also take an
additional reading of each photodiode with all LEDs turned off, in
order to account for any traces of ambient light that might have
penetrated our outer layer. We record all the ensuing 960 signals at a
rate of 60 Hz.

\begin{figure}[t]
  \setlength{\tabcolsep}{0mm}
  \begin{tabular}{c}
    \includegraphics[clip, trim=1.0cm 7.5cm 13.2cm 0cm, height=18mm]{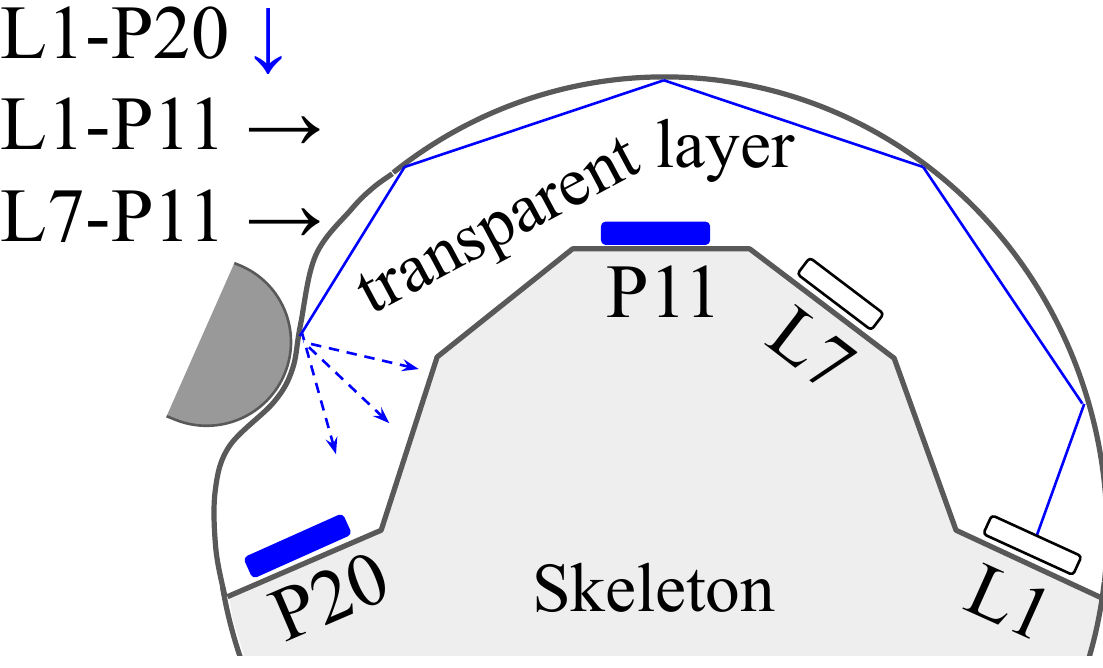}
    \includegraphics[clip, trim=1.1cm 7.5cm 13.2cm 0cm, height=18mm]{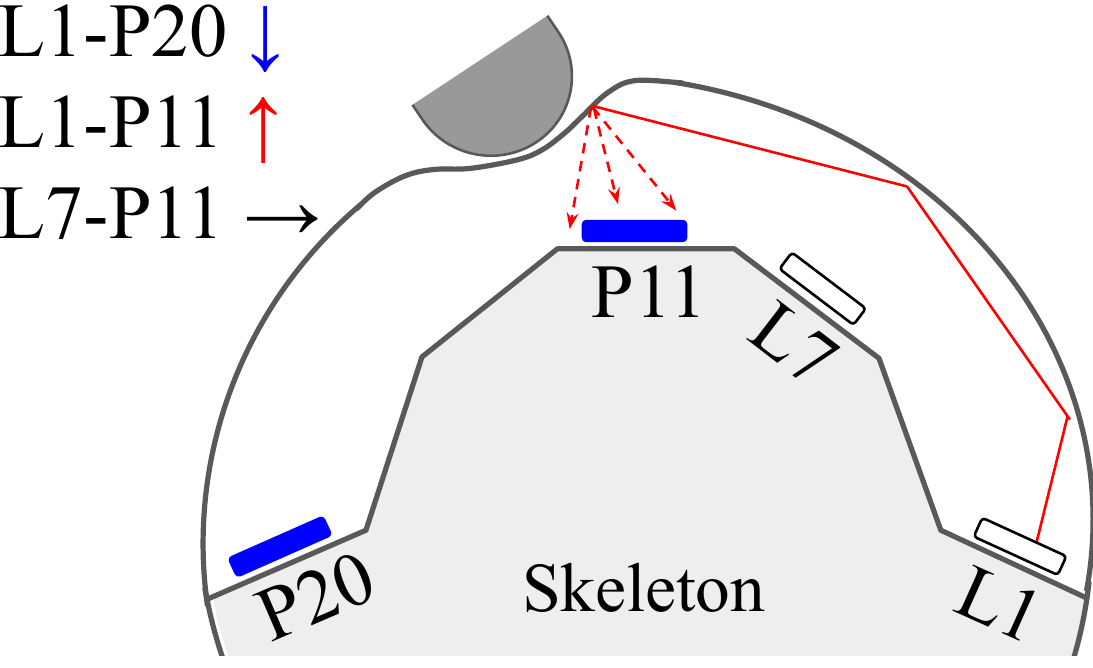}
    \includegraphics[clip, trim=1.2cm 7.5cm 13.2cm 0cm, height=18mm]{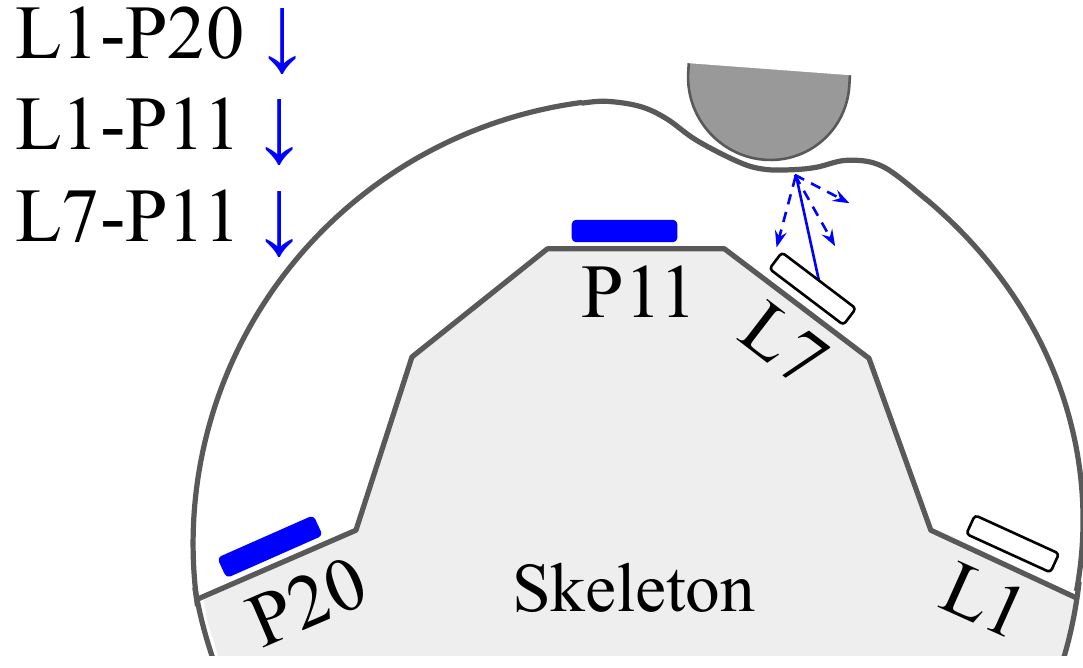}\\[2mm]
  \begin{tabular}{cc}
    \includegraphics[clip, trim=0cm 4.5cm 0cm 5cm, width=0.5\linewidth]{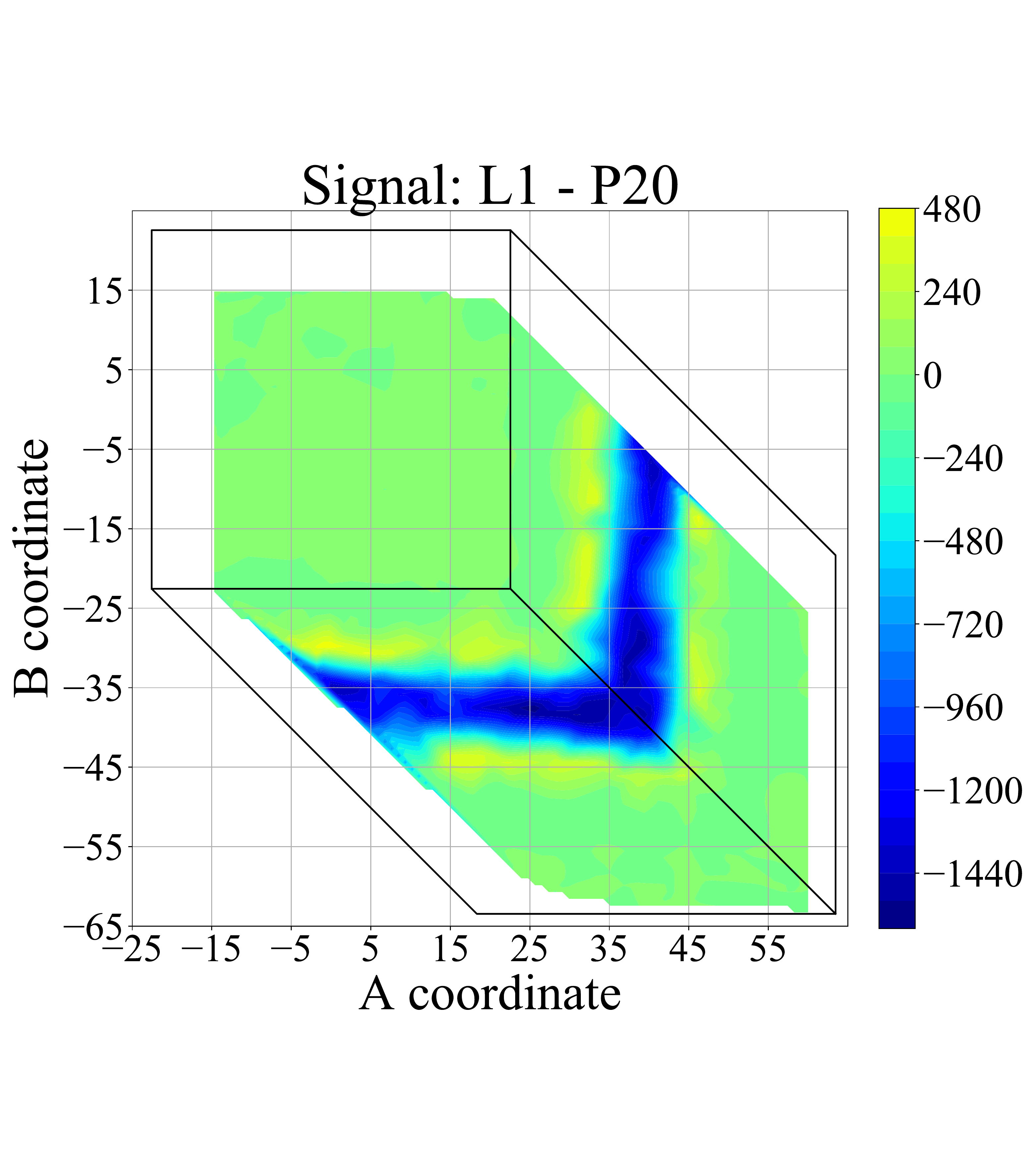}&
    \includegraphics[clip, trim=0cm 4.2cm 0cm 4.7cm, width=0.5\linewidth]{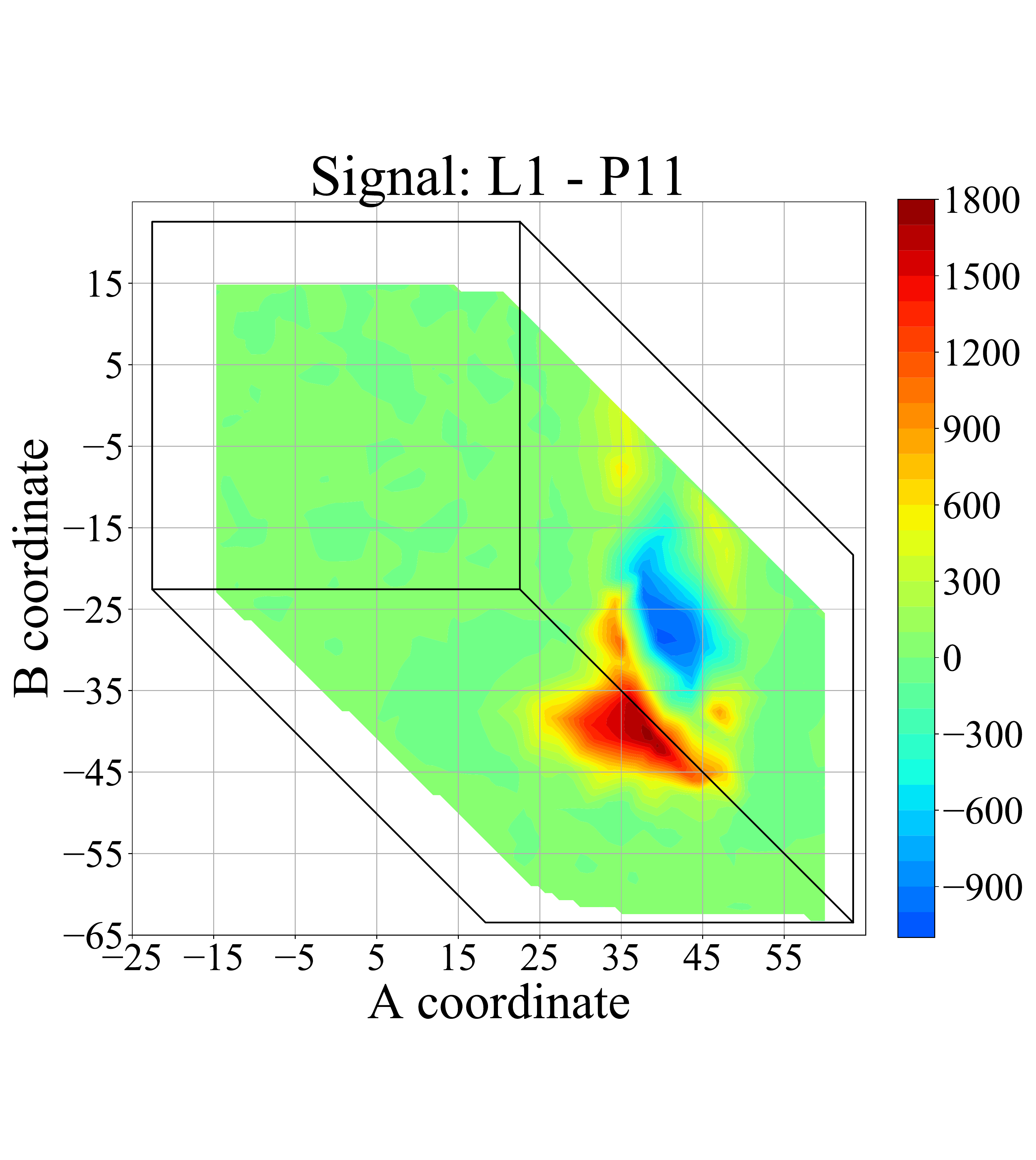}\\
    \includegraphics[clip, trim=0cm 4.5cm 0cm 5cm, width=0.5\linewidth]{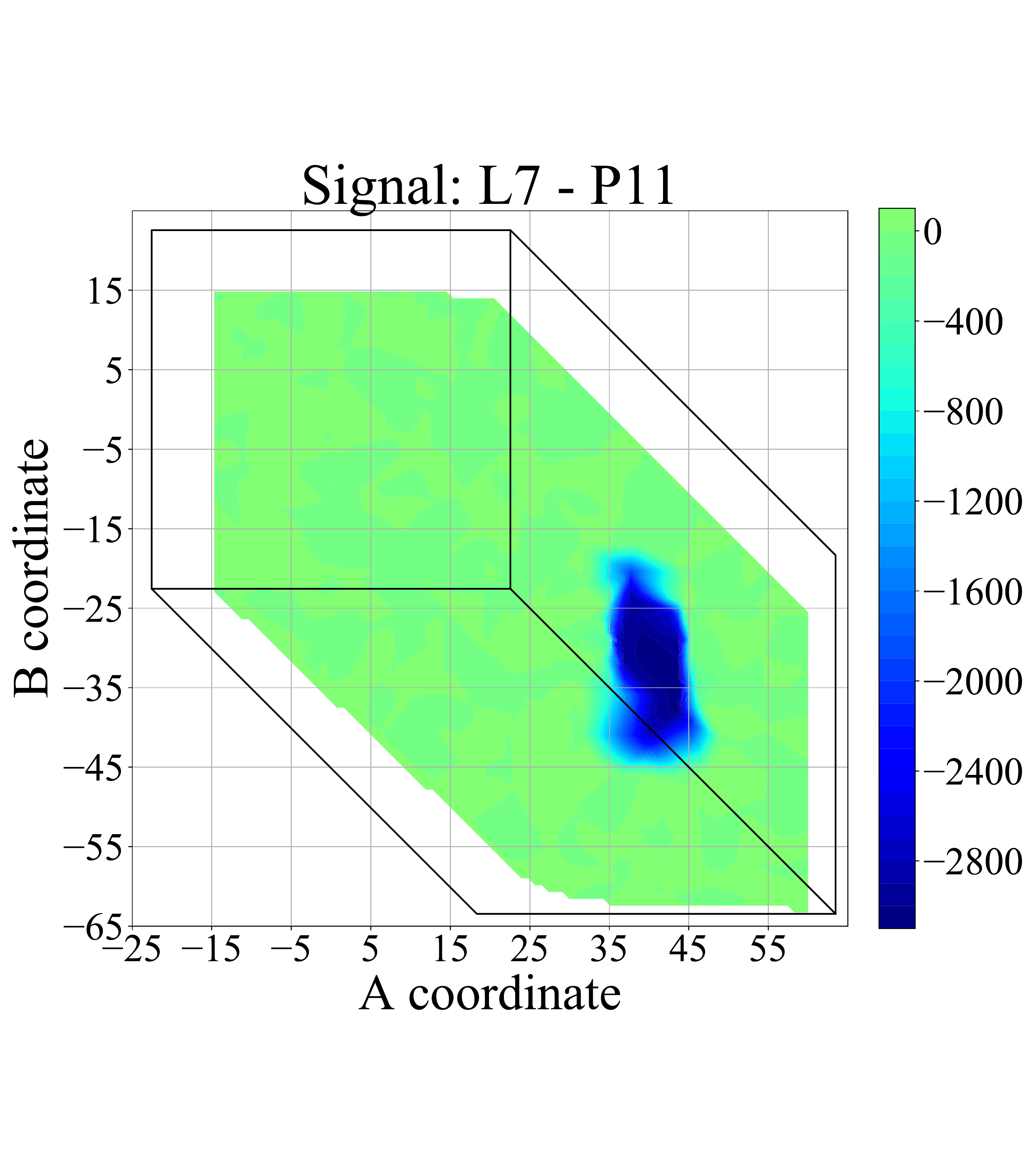}&
    \includegraphics[clip, trim=0cm 4.2cm 0cm 4.7cm, width=0.5\linewidth]{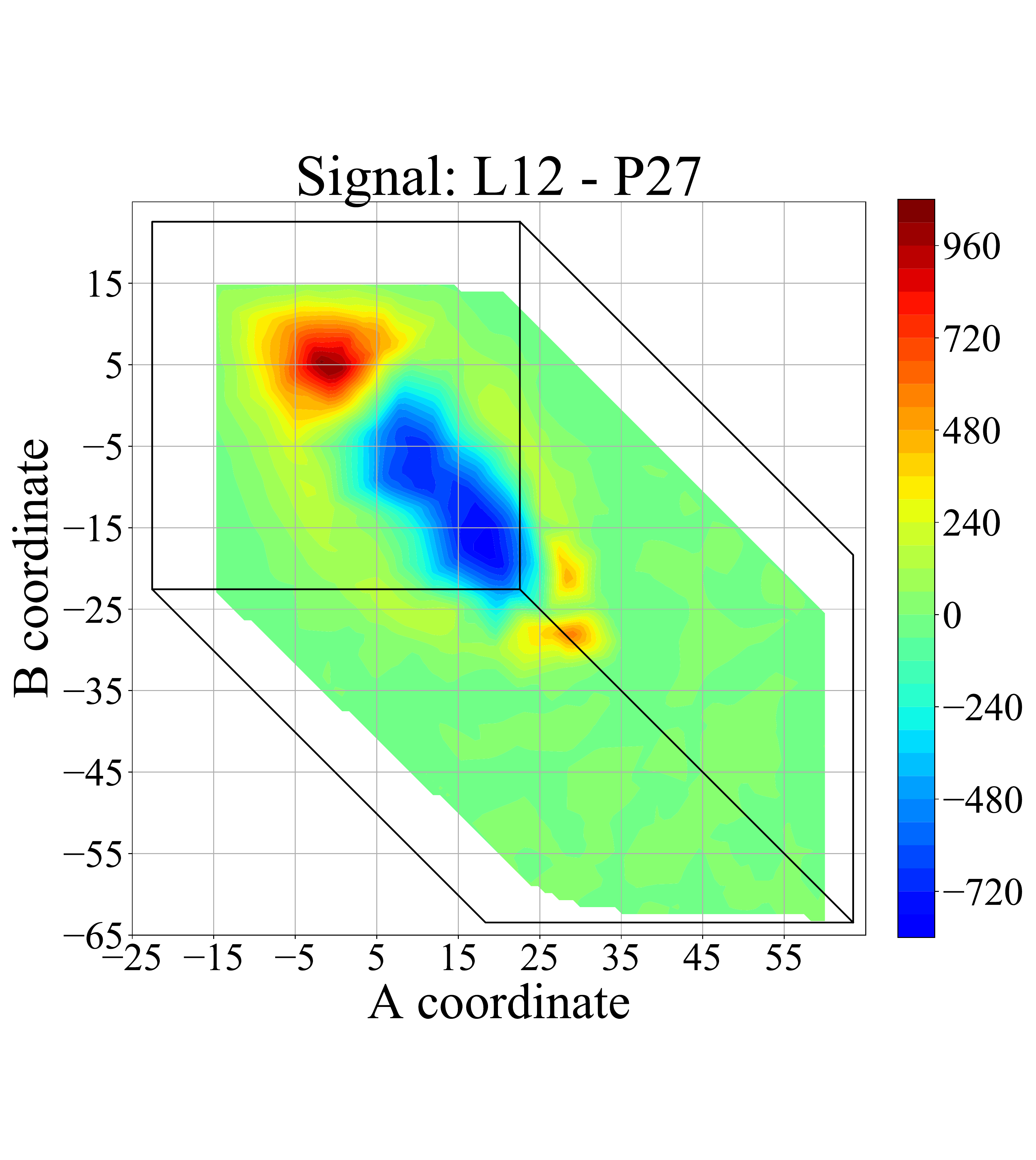}
  \end{tabular}
  \end{tabular}
  \caption{Receptive fields for tactile optical signals. \textbf{Top:}
    illustration of light transport through transparent layer. Each
    LED-photodiode pair provides one signal; indentations cause
    different signals to change in different ways. \textbf{Bottom:}
    Real data from our sensor showing how various signals are affected
    by indentations, based on location (using the $(A,B)$
    parameterization from Fig.~\ref{fig:geometry}). For each signal,
    the heatmap shows the change in raw value caused by an 4 mm deep
    indentation with a hemispherical tip of 10 mm diameter. The
    flexboard locations of all LEDs and diodes referenced here are
    marked on Fig.~\ref{fig:geometry}. }
\label{fig:signals}
\end{figure}

Since we measure light transport between each LED and each photodiode,
we want to ensure that the light emission from each LED is bright
enough to reach distant diodes, but also that it does not saturate
nearby diodes. \modifiedtext{3-4-2}{We use an AD8616 operational
  amplifier on the output of the Teensy digital to analog converter to
  control the brightness for each individual LED. For every LED/diode
  pair, we choose a dedicated LED brightness level such that the
  response of the diode in the undisturbed state of the finger is
  close to the middle point of the diode's output range.} This ensures
that, in the presence of deformation, each of our signals has
``headroom'' to either increase or decrease, depending the position of
the LED and diode relative to the indenter.

\begin{figure}[t!]
  \centering
    \includegraphics[clip, trim=0cm 0cm 0cm 0cm,width=0.9\linewidth]{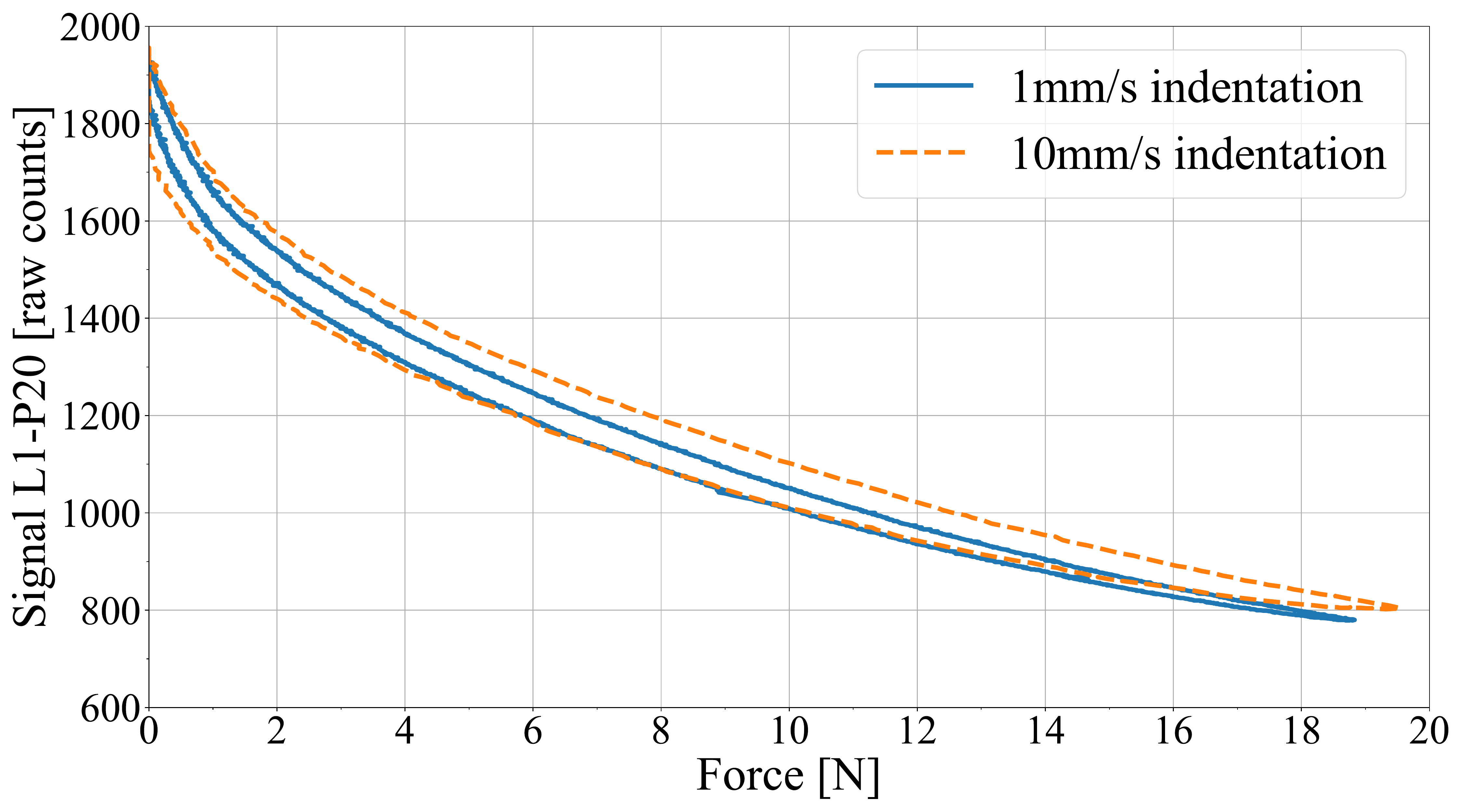}
  \caption{\modifiedtext{1-3}{Hysteresis plot showing two
      indentation-release cycles performed at different speeds. In
      each case, the line with a larger raw signal corresponds to the
      indentation, and the line with a lower raw signal to the
      release.}}
\label{fig:hysteresis}
\end{figure}

Fig.~\ref{fig:signals} illustrates the concept of a receptive field
for a number of representative LED/diode pairs using data collected
from the finger. \modifiedtext{1-3}{Fig.~\ref{fig:hysteresis} also
  shows a representative hysteresis plot, collected for raw signal
  L1-P20 on two indentation-release cycles performed at different
  velocities in the middle of the receptive area for that signal.}

%% In order to evaluate the hysteresis
%% characteristics of our raw data, Supplementary Materials
%% Fig.~\ref{fig:hysteresis} also shows a representative raw signal
%% (L1-P20) over time, as we manually perform three rapid
%% indentation/release cycles in the middle of the receptive area for
%% that signal.

\section{Learning from Tactile Signals}

Our signals are directly correlated with the surface deformation of
the finger. However, building an analytical model to reconstruct the
state of the surface based on the collected signals would be a
daunting task. Thus, this methodology is better fitted to using purely
data-driven algorithms to learn the mapping between this rich set of
signals to various touch parameters of interest. In this study we
focus on determining the location of a contact and the applied normal
force.

\subsection{Data collection} 

\subsubsection{Single touch data}
\label{sec:data_single}
For cases where the finger is contacted in a single location, we
automated data collection with a Universal Robot UR5. We use a custom
end effector mounting for a linear actuator (Physik Instrumente
M-235.5DD) which is fitted with a load cell at its end to record force
measurements (Futek LSB205 10 lbs). At the other end of the load cell
we can mount multiple tips with different geometries.

With this setup, we probe the finger at random locations with
different tips mounted on our load cell. These random locations are
sampled from the defined $(A,B)$ region that represents our sensorized
finger surface. It must be noted that all indentations for this study
are normal to the finger surface. We define a dataset to contain 100
random locations spread throughout the region in $(A,B)$ space, with
the caveat that we enforce a minimum distance of 4 mm between
locations to explore the full surface more homogeneously. (Any new
randomly selected location is rejected if it is closer than 4 mm to
another sample in the same dataset.)

For every location in the $(A,B)$ region we follow the same
procedure. We use the parameterization described above to go from the
$(A,B)$ location to its corresponding Cartesian coordinates, and
compute inverse kinematics for the robot to reach that point in
space. The robot positions the indenting tip normal to the surface at
a distance of 20 mm. At this point we advance the probe until we find
the finger surface using the force measurements from the load
cell. Having established the finger surface, we start logging data from
a depth of -1 mm (negative depths should be interpreted as the tip
being above the surface) and up to 4 mm. (Indentations with the planar
tip only go to a depth of 3 mm to avoid exceeding the load cell
maximum force rating of 10 lbs). \modifiedtext{3-6}{We collect data at
  0.1 mm depth intervals, which yields a total 51 measurements at
  different depths for every location,} hence a dataset with 100
locations constitutes 5100 data points.

\begin{figure*}[t!]
  \centering
  \includegraphics[clip, trim=0.4cm 0.2cm 0.4cm 0.2cm,width=0.33\linewidth]{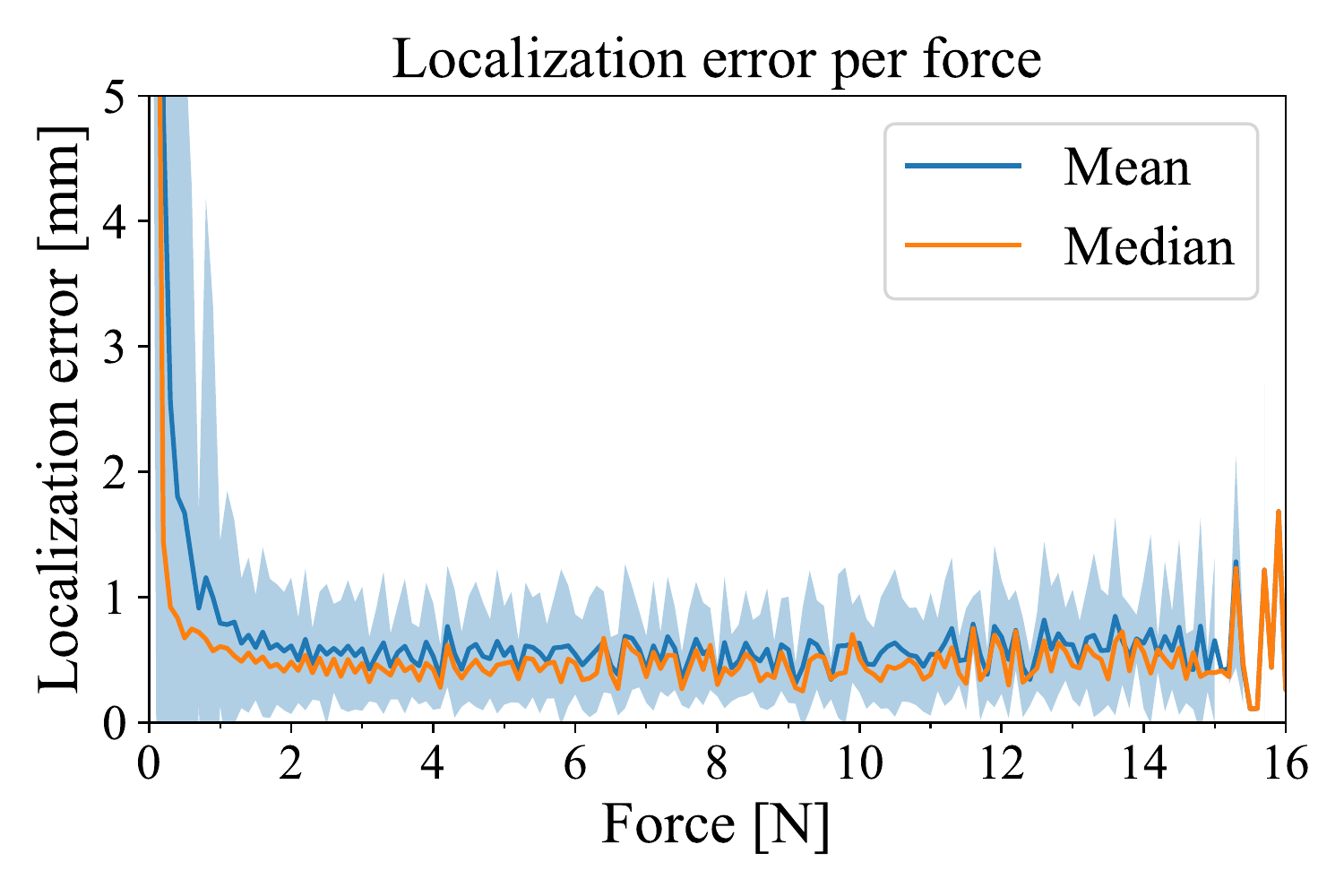}~
  \includegraphics[clip, trim=0.4cm 0.2cm 0.4cm 0.2cm, width=0.33\linewidth]{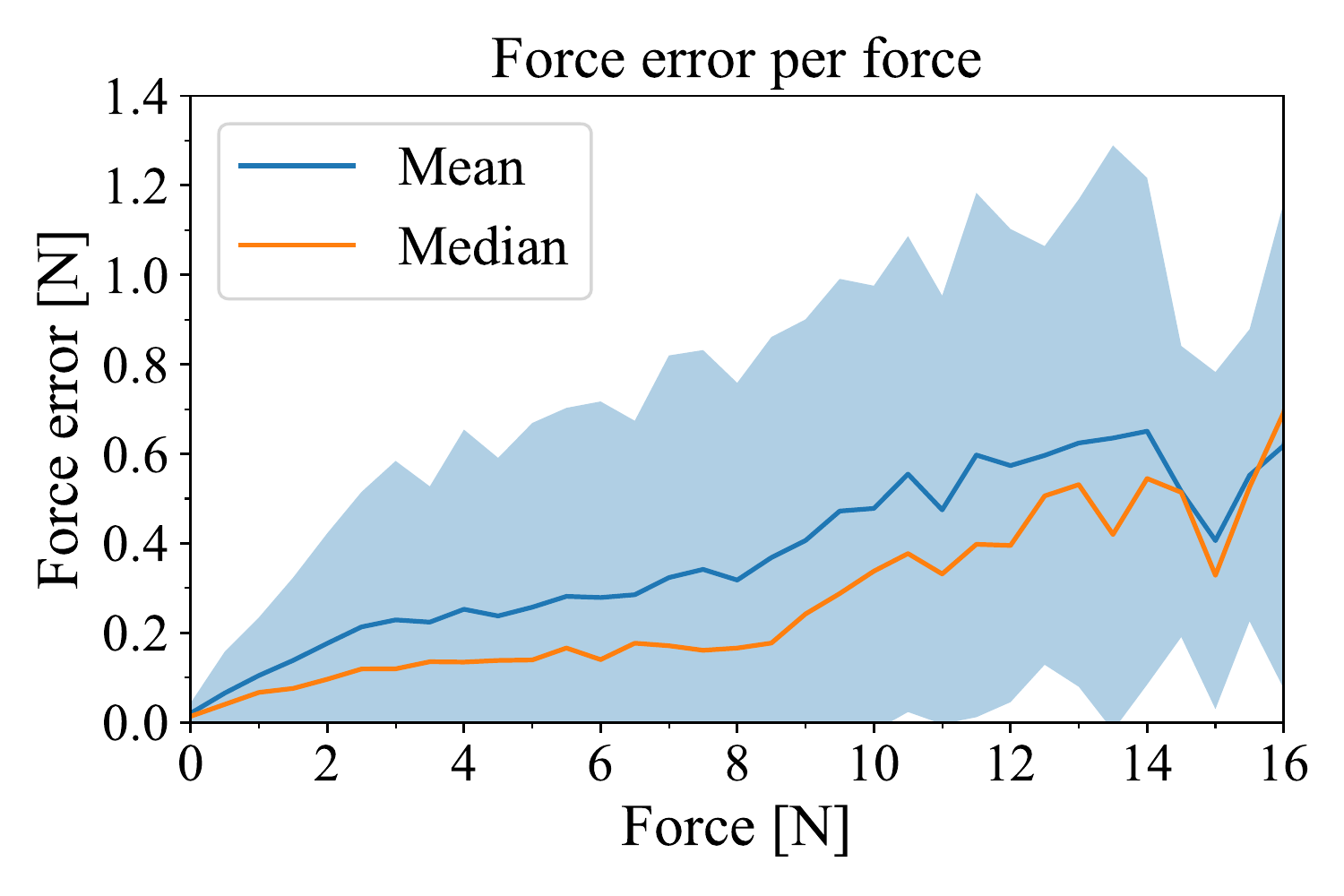}~
  \includegraphics[clip, trim=0.4cm 0.2cm 0.4cm 0.2cm,width=0.33\linewidth]{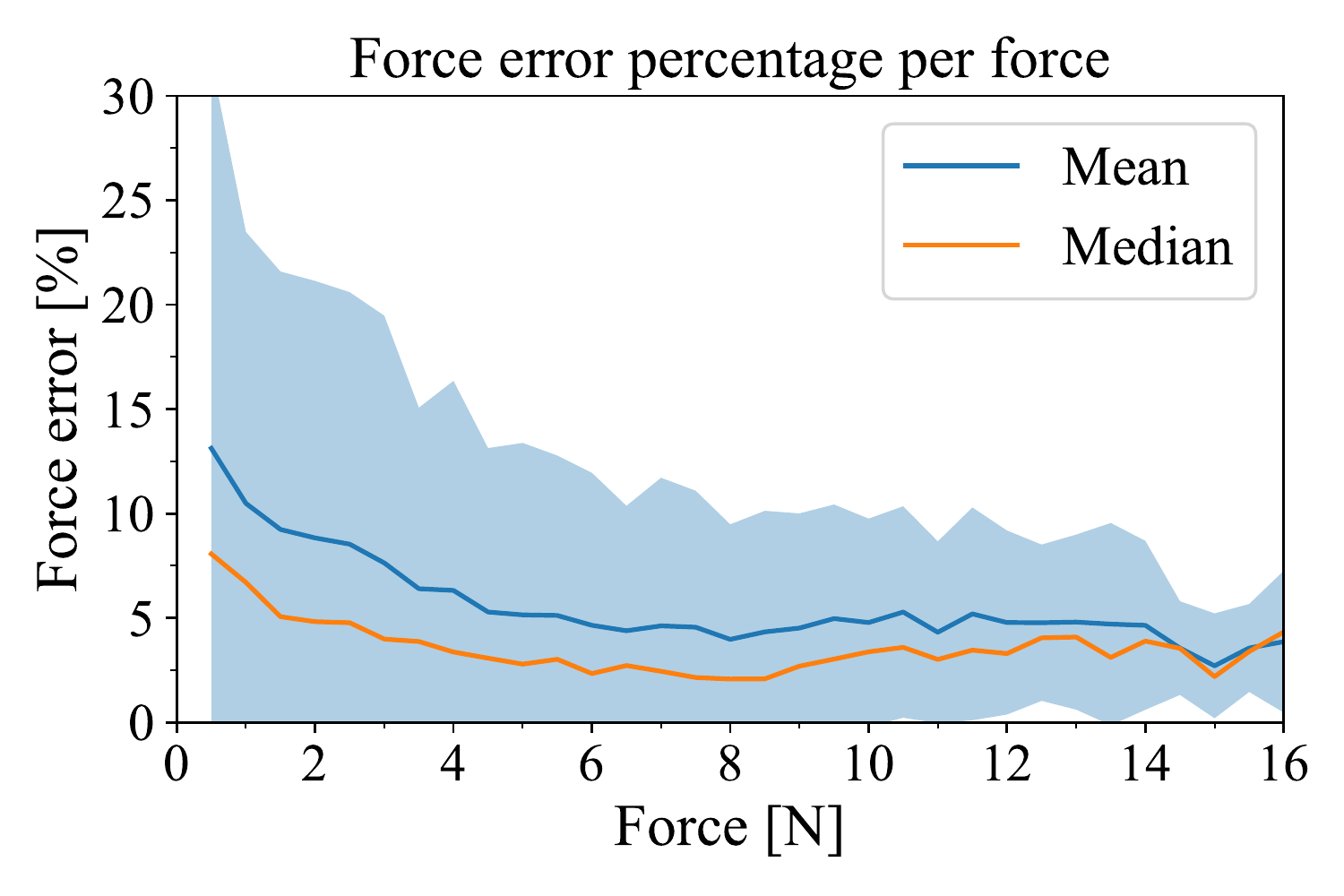}
  \vspace{-7mm}
  \caption{Aggregated localization and contact force prediction
    error. All plots show the mean (blue line),
    \addedtext{3-7}{standard deviation (shaded blue area)} and median
    (orange line) values, computed over all finger locations that we
    have tested. \textbf{Left:} absolute localization error, by
    measured contact normal force. \textbf{Middle:} absolute normal
    force prediction error, by measured contact normal
    force. \textbf{Right: } relative normal force prediction error as
    a percentage of measured normal force, by measured normal force.}
  \label{fig:aggregate}
  \vspace{-3mm}
\end{figure*}

Each measurement $i$ results in a tuple of the form
$\Phi_i^{single}=(a_i,b_i,d_i,f_i,tip,r_1,..,r_{990})$ where $(a_i, b_i)$ is
the indentation location in $AB$ space, $d_i$ is the depth at which
the measurement was taken, $f_i$ is the measured force in newtons,
$tip$ is the id corresponding to the indenter geometry used and
$(r_1,..,r_{990})$ is the feature vector with dimension 990 containing
all light measurements between all 32 LEDs and 30 photodiodes and an
additional 30 measurements which correspond to turning off all LEDs
and measuring any traces of ambient light received by each photodiode,
which is then subtracted from the main 960 signals.

\addedtext{3-3}{Using this dataset, we checked how many of these raw
  signals carry useful information. We consider a signal as useful if,
  in response to an indentation anywhere on the finger surface, the
  signal changes by more than three times its own standard deviation
  in the undisturbed state (used as a measure of noise). We found that
  917 our of 960 signals exceed this threshold, confirming the
  hypothesis that our overlapping signals approach leads to a large
  number of useful signals.}

\subsubsection{Multi-touch data}
\label{sec:data_multi}
\modifiedtext{2-3}{To collect data where the finger is potentially touched in multiple
locations, we revert to a manual procedure. We divide the cylindrical
part of the finger into a 4$\times$5 regular grid. For one data point, a
human experimenter contacts the finger in either a single cell, or
simultaneously in two cells, chosen at random. Contact is made in all
cases with hemispherical indenters (10 mm diameter). Each measurement
$i$ thus results in a tuple of the form
$\Phi_i^{multi}=(c_1,..,c_{20},r_1,..,r_{990})$, where $c_j$ is a
binary signal indicating if cell $j$ is being contacted, and the
feature vector $(r_1,..,r_{990})$ comprises tactile finger signals as
above. We note that contact forces are not measured in this case.}
 
\subsection{Learning algorithms}

For all experiments on our finger, we use feed forward neural network
architectures. The architectures are slightly customized to support
the related task. First, we describe the architecture for contact
localization and force detection, followed by the neural network for
multi-touch detection.

For localization and force prediction, we use a multi-task neural
network with five hidden layers. Each hidden layer uses batch
normalization and the ReLU activation function. The first three hidden
layers are shared between both tasks, with 512, 256 and 128 activation
units respectively. Afterwards, each output has two individual hidden
layers with 64 and 32 hidden units respectively. We use mean squared
error as a loss function. The network is trained for 600 epochs with
ADAMoptimizer, a batch size of 128 and an initial learning rate of
0.001. After 500 epochs the learning rate is decreased to 0.0001.  For
localization prediction, the training dataset is filtered to only use
positive depth, because the network cannot be expected to predict
location when touch is not occurring.

Collecting precise multi touch data would require two robotic arms
operating on the same finger without colliding with each other,
simultaneously. In the absence of such a setup, we used a simpler
manual procedure. We divided a section of the finger into a 4$\times$5 grid
cell (in total 20 cells) and collected data via manual
indentation. The training process randomly selects one or two out of
the 20 cells and shows their identity to the experimenter. The
experimenter then indents the respective cells, and records the
ensuing tactile data. We note that exact contact force is unavailable
when using this procedure.

The resulting dataset is one order of magnitude smaller than the ones
obtained with our automated process and, therefore, a simpler neural
network is required. For multi-touch prediction, a two hidden layer
feed forward neural architecture is used with 128 and 32 hidden units
respectively. The hidden layers use the ReLU as activation function,
but no batch normalization. The output consists of 20 independent
binary predictions, each indicating the probability of touch for a
certain cell. For this task, we used the sigmoid cross entropy
loss. The network was trained for 400 epochs with ADAMoptimizer with a
batch size of 128 and initial learning rate of 0.001, which was
reduced to 0.0001 after 200 epochs.

\section{Results}

\subsection{Touch localization and force detection}

We first quantify the ability of our finger to predict the location of
touch. We perform this test on a corpus of single-touch data comprising 4,896
indentations at 96 different locations and with different forces,
collected using the procedure outlined
in~\ref{sec:data_single}. Fig.~\ref{fig:aggregate} shows the
localization error at different force levels over the entire test set
(mean and median). Note that while our prediction is made in the
dimensionless $(A,B)$ space, we first convert both the prediction and
the ground truth location back to Cartesian $(x,y,z)$ space in order
to calculate the error in distance units (mm). We also note that the
smallest force that our load cell can accurately detect is 0.2 N, and
any force measurement below that value indicates either no contact, or
very slight touch.

We observe that our finger quickly achieves sub-millimeter accuracy in
touch localization, once the contact force exceeds 0.2 N. The error
continues to decrease until a force level of approximately 1 N, at which
point performance stabilizes throughout the rest of the range of
forces we tested. At 0 contact force (no touch), the localization
error is 30.3 mm (median), in line with a random chance guess.

We now turn our attention to the ability to also predict normal
contact force. Using the same datasets as before, we train a neural
network regressor to predict contact normal force, using load cell
data as ground truth. The absolute error (difference between
predicted and real force), as well as the error as a percentage of the
ground truth applied force, are both shown in Fig.~\ref{fig:aggregate},
for different contact forces. 

We note that the relative error is below 10\% (median) even for very
light touches, and reduces to as low as 2\% (median) in the middle of
our range (around 8 N). At that point, relative error stabilizes
around 3\% (median) and 5\% (mean) for the rest of the range, up to 16
N. In absolute terms, this means that while error grows together with
the applied force, it does so slowly, and, for a force range between 2
and 9 N, \modifiedtext{3-8-1}{the median error} is approximately
constant, and below 0.2 N.

\begin{figure}[t!]
  \centering
    \setlength{\tabcolsep}{1mm}
  \begin{tabular}{cc}  
    \includegraphics[clip, trim=0.3cm 1.4cm 0.3cm 1.0cm,width=0.45\linewidth]{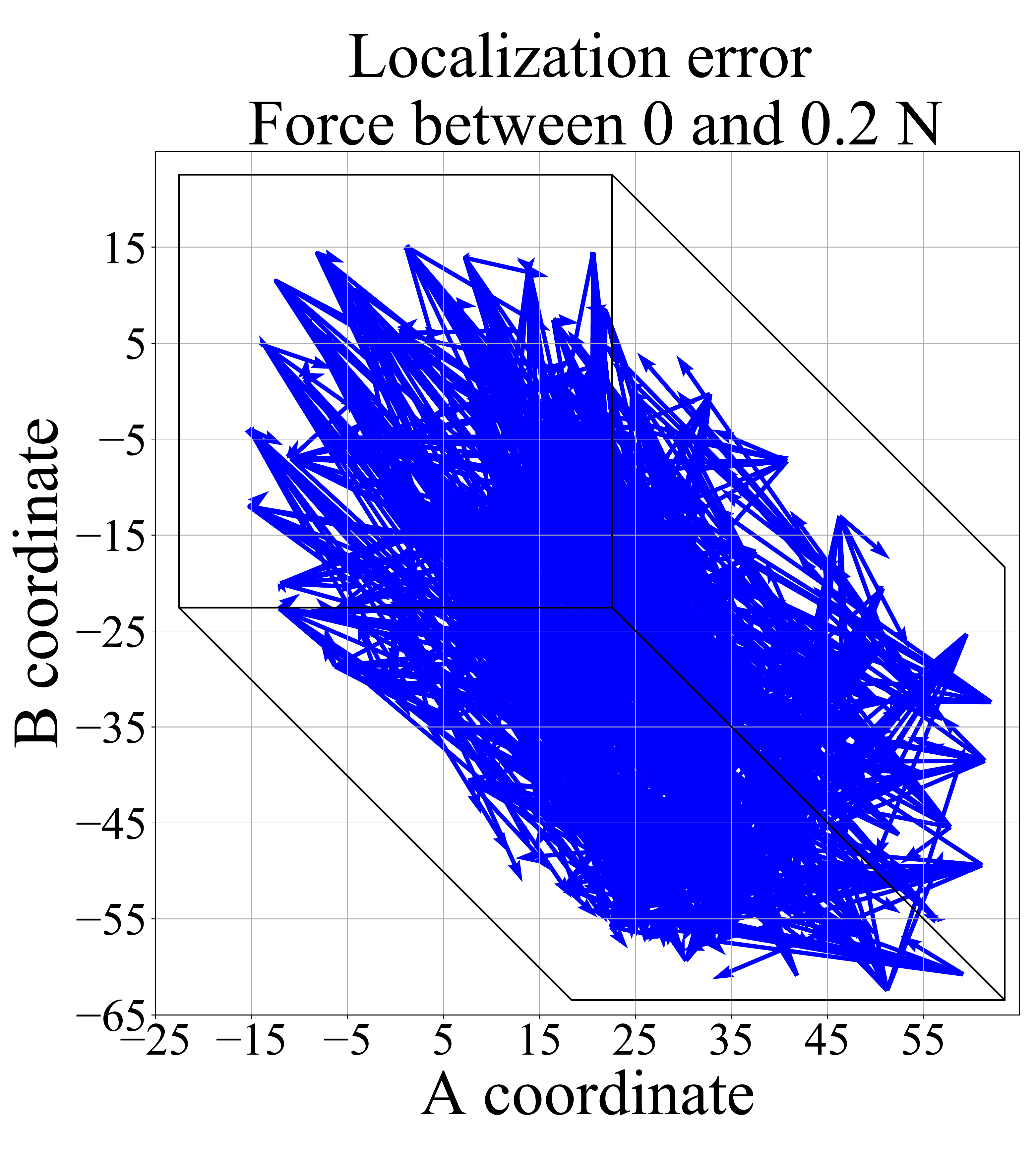}&
    \includegraphics[clip, trim=0.3cm 1.4cm 0.3cm 1.0cm,width=0.45\linewidth]{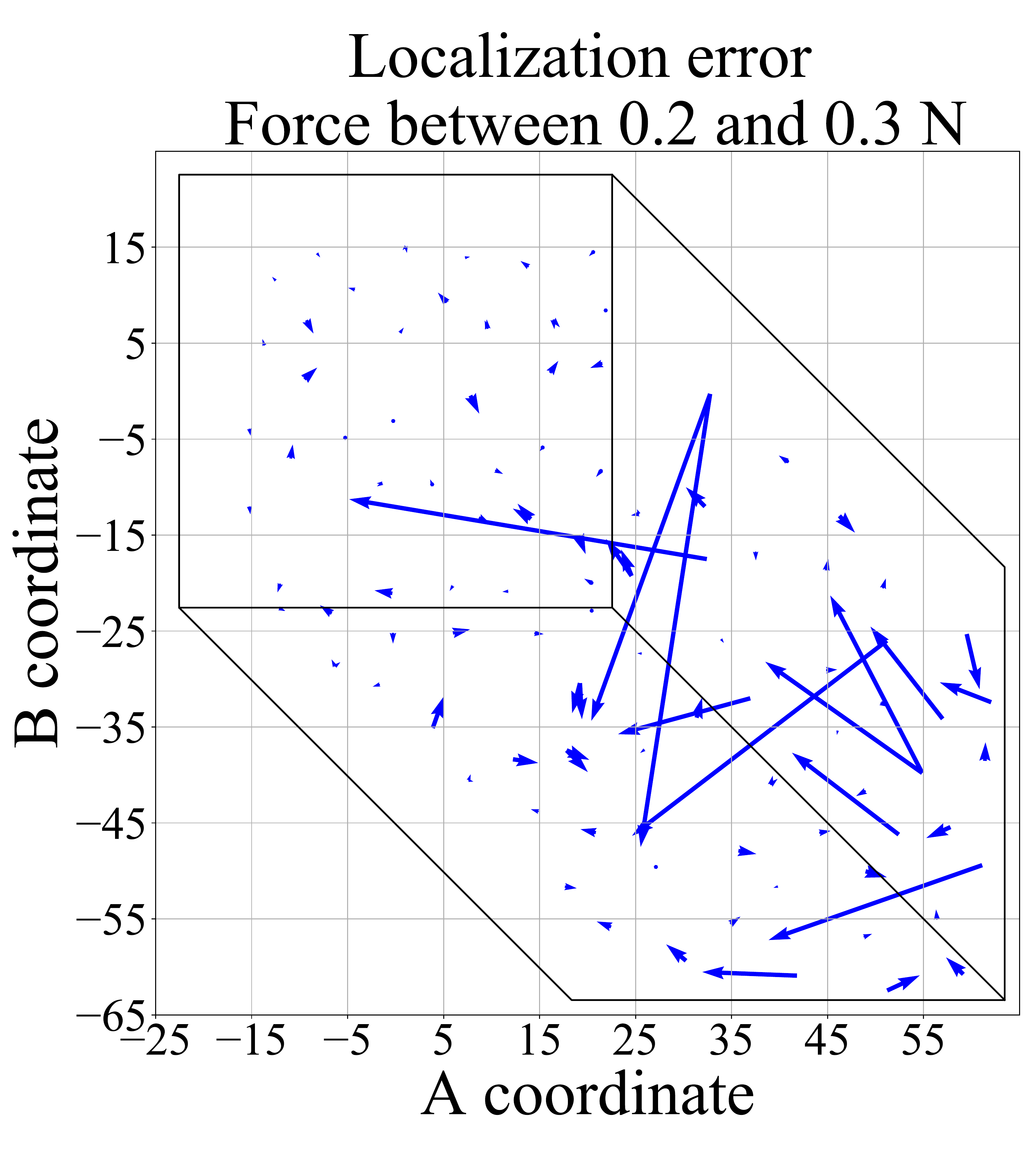}\\[2mm]
    \includegraphics[clip, trim=0.3cm 1.4cm 0.3cm 1.0cm,width=0.45\linewidth]{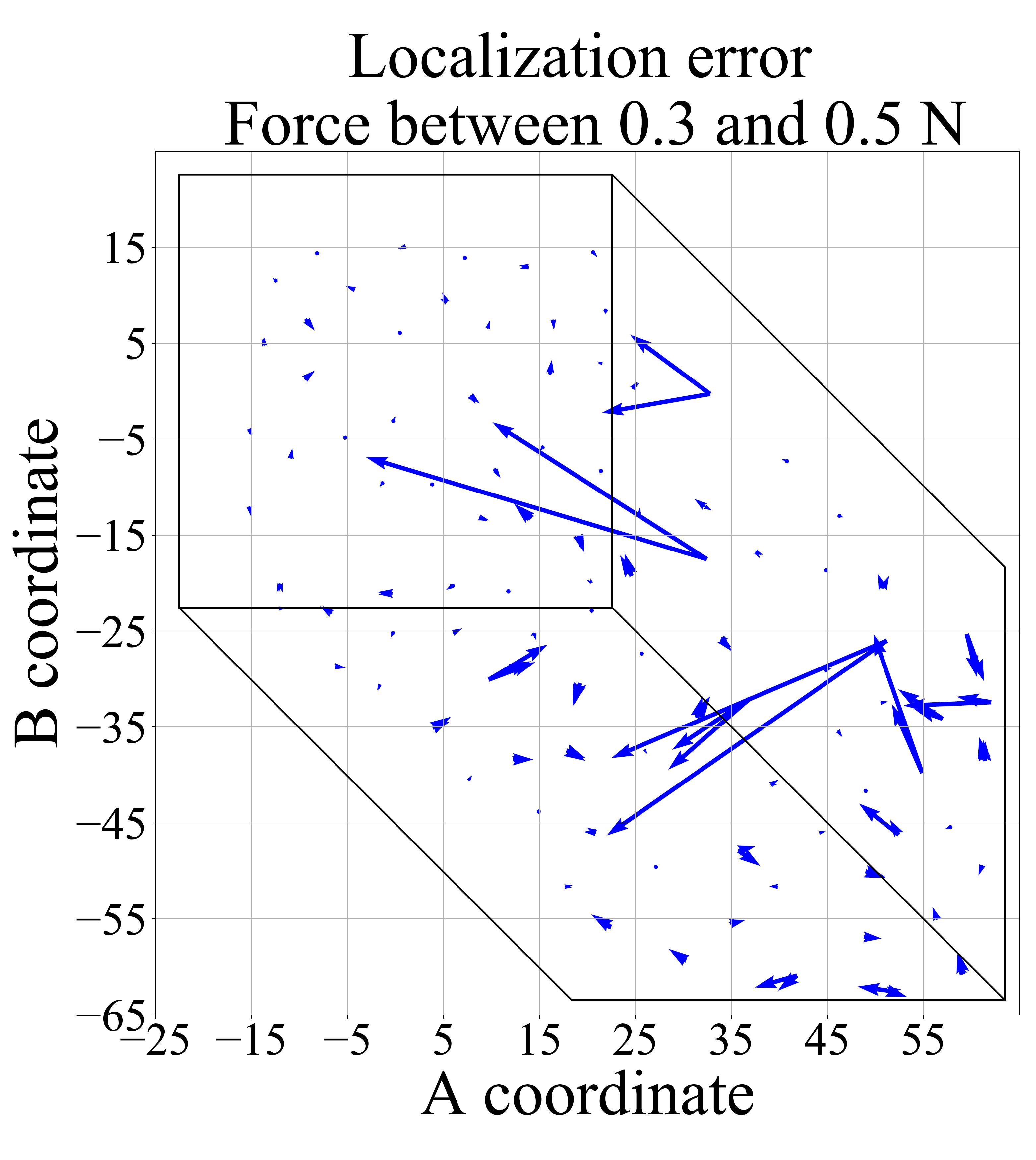}&
    \includegraphics[clip, trim=0.3cm 1.4cm 0.3cm 1.0cm,width=0.45\linewidth]{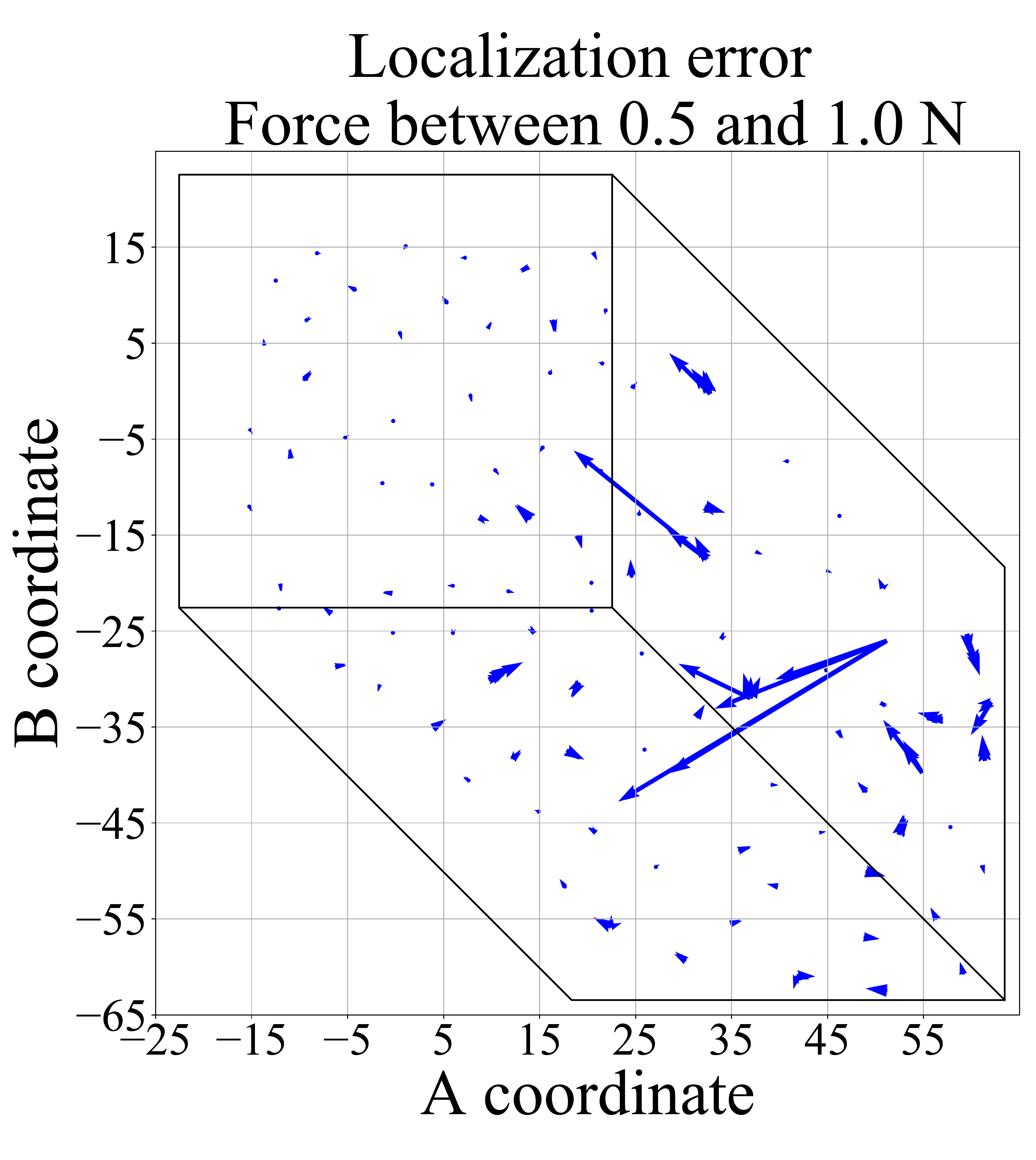}\\[2mm]
    \includegraphics[clip, trim=0.3cm 1.4cm 0.3cm 1.0cm,width=0.45\linewidth]{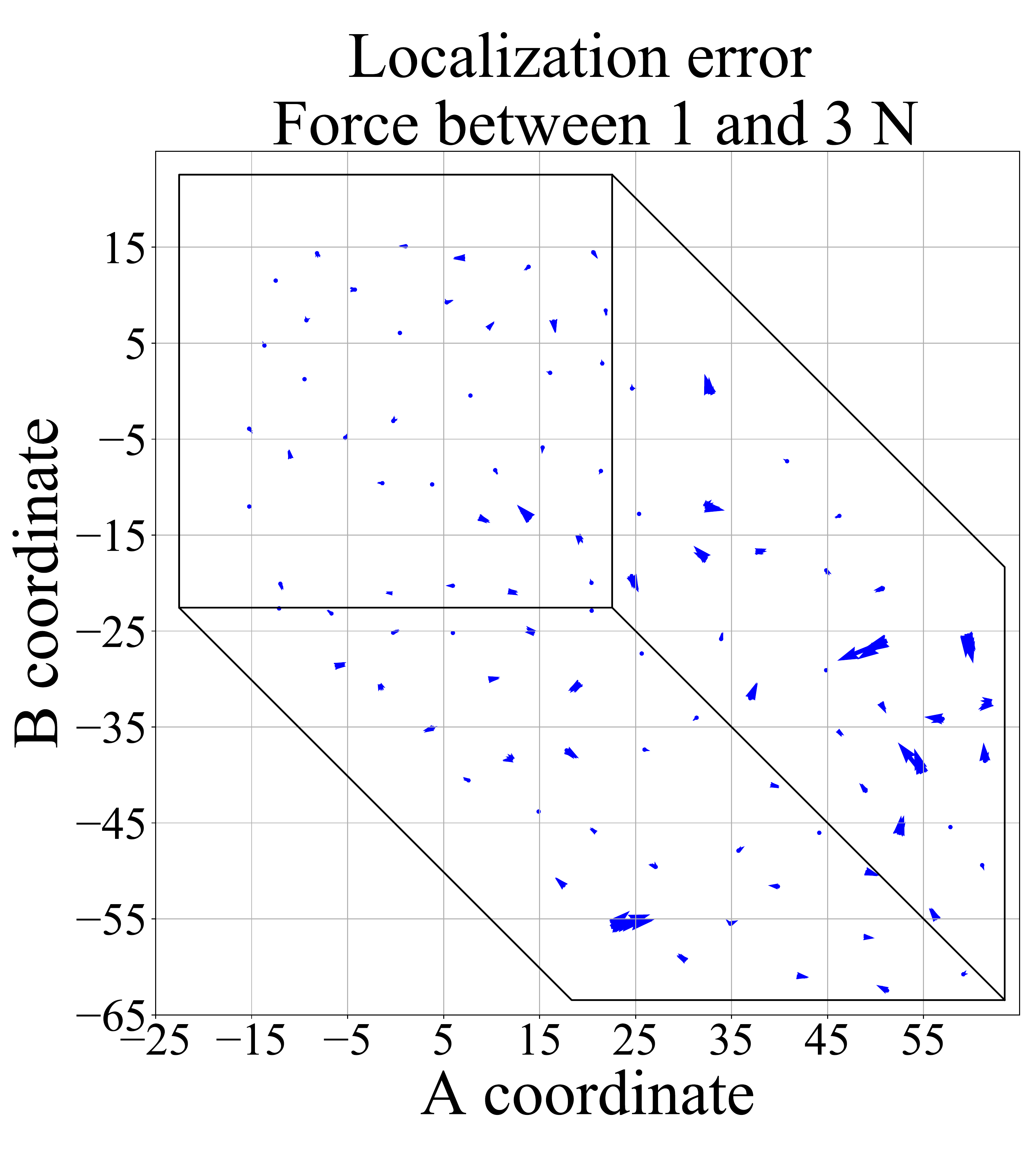}&
    \includegraphics[clip, trim=0.3cm 1.4cm 0.3cm 1.0cm,width=0.45\linewidth]{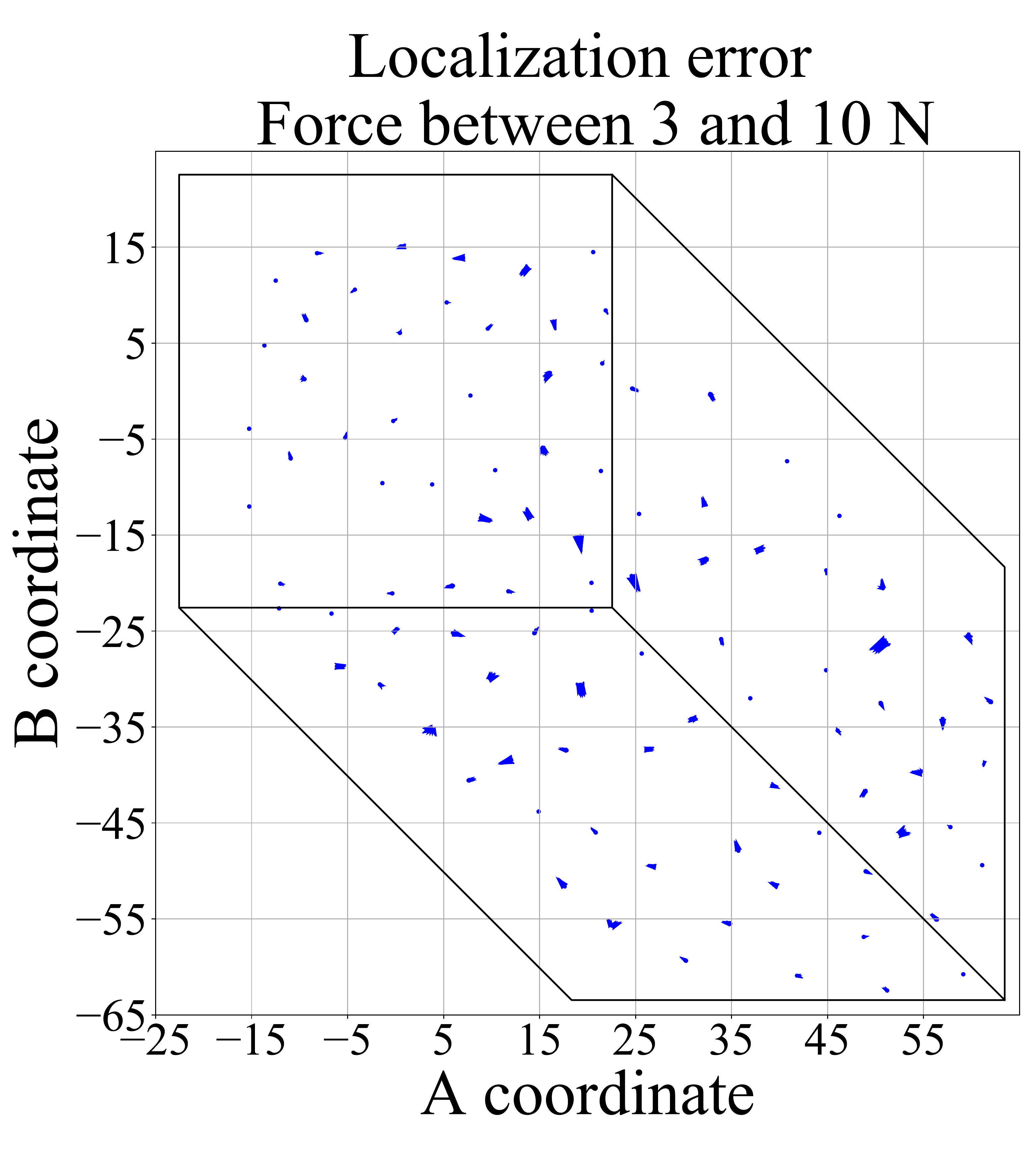}
 \end{tabular}
  \caption{Localization prediction error for each test point, shown in
    the $(A,B)$ space illustrated in Fig.~\ref{fig:geometry}. For each
    test point, we render an arrow; the base of the arrow shows the
    true location of touch, while the tip of the arrow shows the
    predicted location; a shorter arrow thus corresponds to lower
    error. We bin test points together based on applied force, with
    each plot corresponding to data from one bin, as indicated in the
    plot title.}
  \label{fig:locations}
  \vspace{-4mm}
\end{figure}

For more insights into how localization performance varies across its
surface, we would like to study if the magnitude or direction of the
error is affected by the area of the finger that is being
touched. Fig.~\ref{fig:locations} shows this error for every data
point, grouped by force levels. For clarity, the data is shown in the
two-dimensional $(A,B)$ space, as illustrated in
Fig.~\ref{fig:geometry} and formally defined in the Appendix. We
notice that at no contact and below a force level we can reliably
detect (0 to 0.2 N), errors are in line with random guesses. At low
forces (0.2 to 0.5 N) most of the functional area of the finger has
good accuracy, with the errors occurring along the edges of the
cylindrical part. For forces beyond 1 N, larger errors have been
eliminated throughout the functional area.

\subsection{Robustness to indenter shape}

The results shown so far have been obtained by using a single indenter
tip. What happens if the shape of the indenter varies? Furthermore,
what is the performance level if, at test time, the finger makes
contact with an indenter different from the one(s) using during
training? These analyses would be indicative of performance in complex
environments, where the robot might interact with objects of varying,
and potentially unknown shape, and in various configurations.

To evaluate this performance, we collected data using multiple
indenters. In addition to the 10 mm diameter hemispherical tip used so
far, we added a planar tip (circular with a 15 mm radius), a sharp
corner, and an edge used in two different orientations (horizontal and
vertical), for a total of five different indentation geometries. We
used this dataset to test localization performance with two different
approaches.

\begin{figure}[t]
  \centering
  \setlength{\tabcolsep}{2mm}
  \begin{tabular}{cc}
    \includegraphics[clip,
      trim=6.5cm 0cm 0cm 0cm,height=0.4\linewidth]{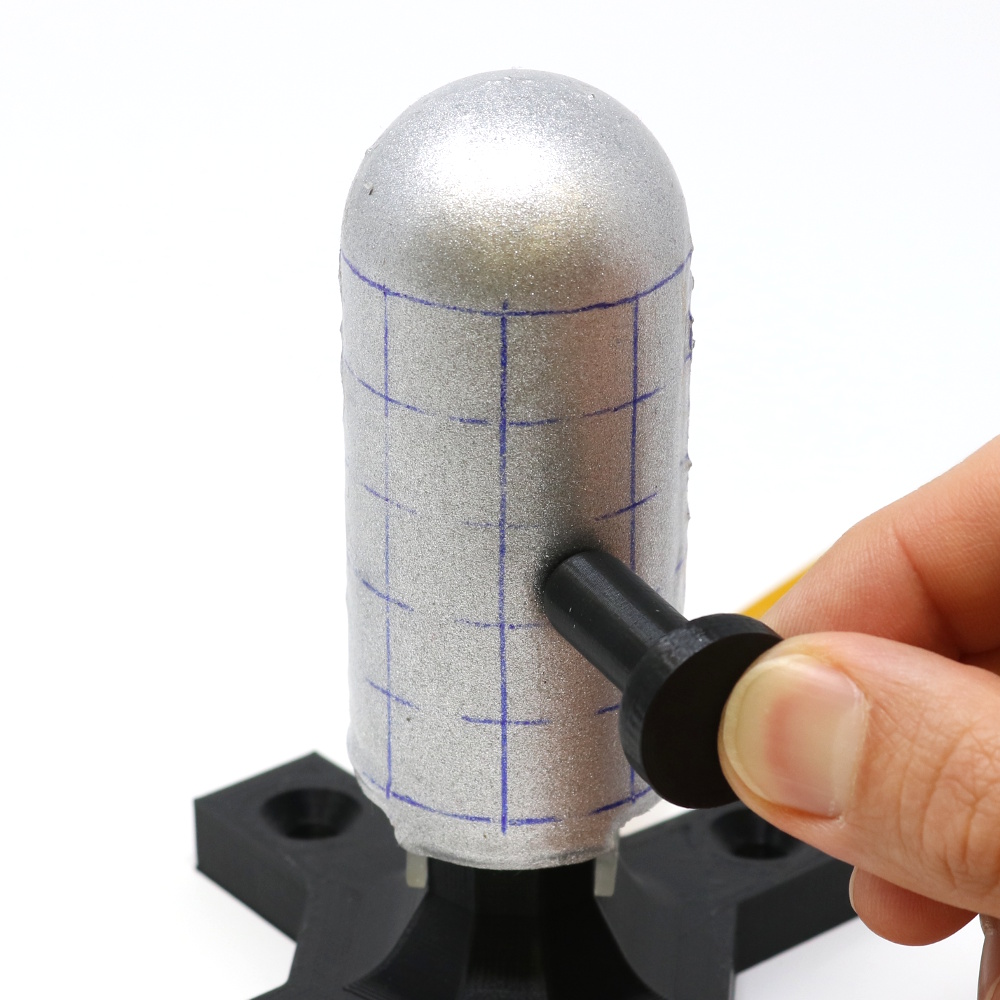}&
    \includegraphics[height=0.4\linewidth]{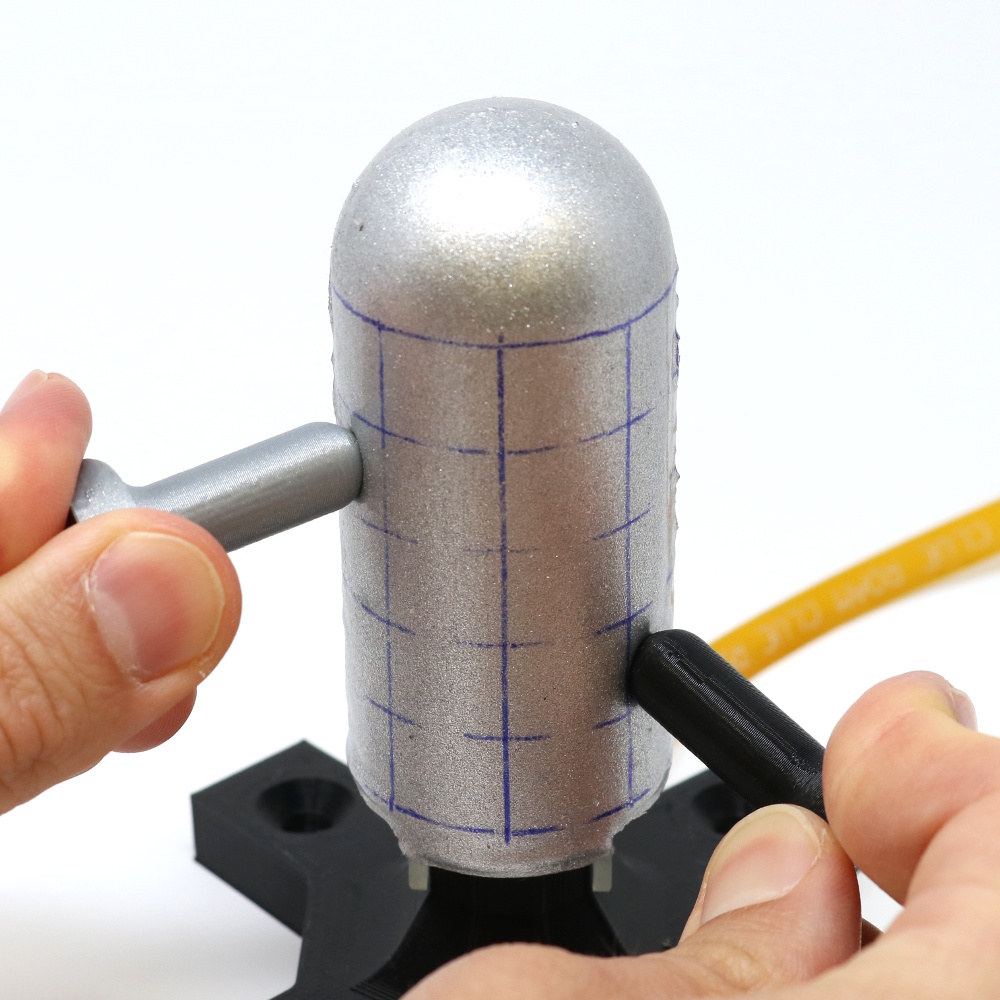}
  \end{tabular}
  \setlength{\tabcolsep}{0mm}
  \begin{tabular}{cccc}
    \includegraphics[clip, trim=4.5cm 0cm 4cm 0cm,height=30mm]{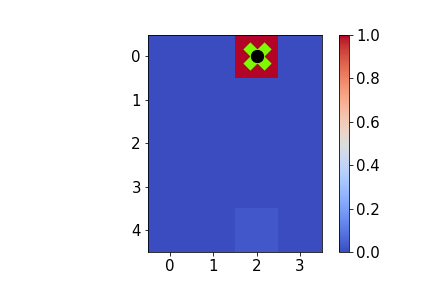}&
    \includegraphics[clip, trim=4.5cm 0cm 4cm 0cm,height=30mm]{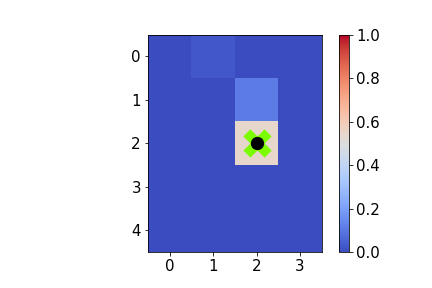}&
    \includegraphics[clip, trim=4.5cm 0cm 4cm 0cm,height=30mm]{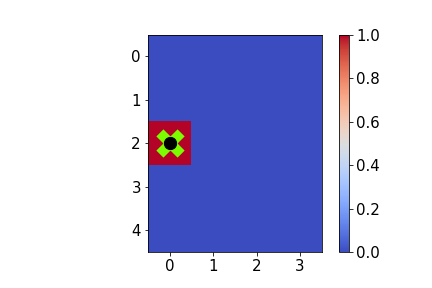}&
    \includegraphics[clip, trim=4.5cm 0cm 1cm 0cm,height=30mm]{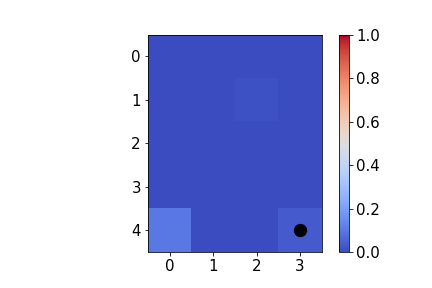}\\[-2mm]
    \includegraphics[clip, trim=4.5cm 0cm 4cm 0cm,height=30mm]{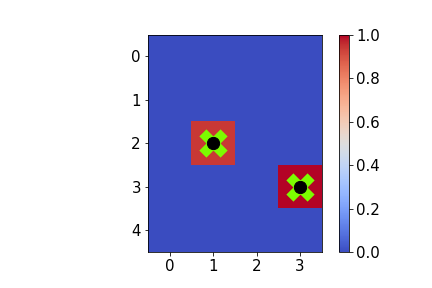}&
    \includegraphics[clip, trim=4.5cm 0cm 4cm 0cm,height=30mm]{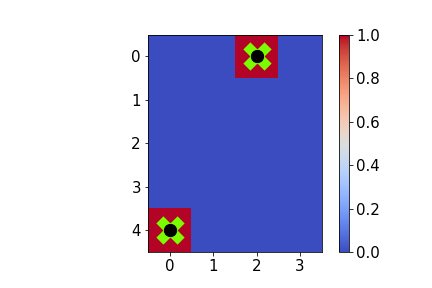}&
    \includegraphics[clip, trim=4.5cm 0cm 4cm 0cm,height=30mm]{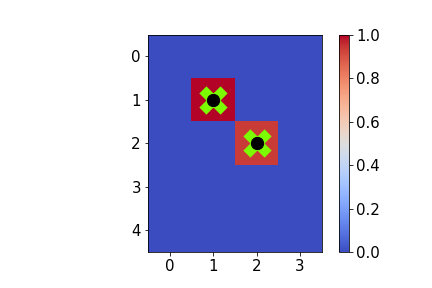}&
    \includegraphics[clip, trim=4.5cm 0cm 1cm 0cm,height=30mm]{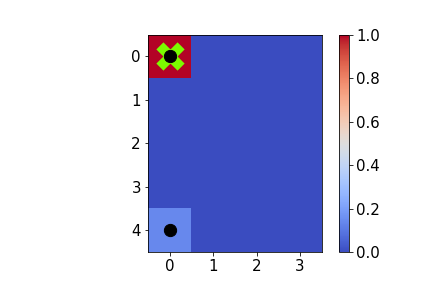}
 \end{tabular}
  \caption{Multitouch prediction results.\textbf{Top:} finger with a
    4$\times$5 discrete grid marked on cylindrical area, and experimenter
    manually indenting one or two cells simultaneously; our model aims
    to predict the number and identity of indented
    cells. \textbf{Bottom:} examples of multitouch predictions from
    our model. Each image shows one test case; black dots indicate
    ground truth identity of indented cells. The color of each cell
    indicates the predicted probability of touch in that respective
    cell, using the colormap shown in the right. Cells where the
    predicted touch probability exceeds 0.5 are marked by a green
    cross. Rightmost images in each row are examples of incorrect
    classifications.}
  \label{fig:multitouch}
\end{figure}

\begin{table*}[t!]
\small
\centering
% \caption{Localization performance with multiple indenter
%     shapes. For each indenter, we show localization error for two
%   models: one trained without data from the indenter being tested
%   (``Leave-one-out'') and one trained with data from all indenters
%   (``All inclusive''). We report performance aggregated over five
%   force intervals; for each force interval, we report the mean and
%   median localization error. Last two rows show performance averaged
%   over all indenter shapes.}  

\caption{For each indenter shape we show localization error for two
  models: one trained without data from the indenter being tested
  (``Leave-one-out'') and one trained with data from all indenters
  (``All inclusive'').}  

\setlength{\tabcolsep}{0.9mm}
\vspace{0mm}
\begin{tabular}{cc|cc|cc|cc|cc|cc}
\multirow{4}{*}{\textbf{Tip}} &
\multirow{4}{*}{\textbf{Model}} &
\multicolumn{2}{c|}{Error (mm)} &
\multicolumn{2}{c|}{Error (mm)} &
\multicolumn{2}{c|}{Error (mm)} &
\multicolumn{2}{c|}{Error (mm)} &
\multicolumn{2}{c}{Error (mm)} \\
&
&
\multicolumn{2}{c|}{for forces} &
\multicolumn{2}{c|}{for forces} &
\multicolumn{2}{c|}{for forces} &
\multicolumn{2}{c|}{for forces} &
\multicolumn{2}{c}{for forces} \\
&
&
\multicolumn{2}{c|}{0.2-0.3 N} &
\multicolumn{2}{c|}{0.3-0.5 N} &
\multicolumn{2}{c|}{0.5-1.0 N} &
\multicolumn{2}{c|}{1.0-3.0 N} &
\multicolumn{2}{c}{3.0-10 N} \\
&
&
Mean&
Median&
Mean&
Median&
Mean&
Median&
Mean&
Median&
Mean&
Median
\\\hline
\\[-3mm]\hline
\multirow{2}{*}{Edge (H)}        & Leave-one-out & 13.0 & 7.5  & 8.3 & 4.0 & 4.2 & 2.5 & 2.5 & 1.7 & 3.7 & 2.1 \\
                                 & All inclusive & 7.8  & 2.9  & 3.2 & 1.9 & 1.6 & 1.2 & 1.0 & 0.9 & 0.8 & 0.7 \\\hline
\multirow{2}{*}{Edge (V)}        & Leave-one-out & 12.6 & 13.0 & 6.8 & 6.3 & 3.8 & 3.3 & 3.0 & 2.9 & 1.8 & 1.7 \\
                                 & All inclusive & 2.8  & 3.6  & 2.7 & 3.2 & 1.6 & 1.6 & 1.1 & 1.1 & 0.8 & 0.9 \\\hline
\multirow{2}{*}{Planar}          & Leave-one-out & 7.5  & 3.3  & 5.6 & 2.7 & 2.8 & 2.0 & 1.9 & 1.5 & 1.5 & 1.3 \\
                                 & All inclusive & 1.2  & 0.9  & 0.9 & 0.8 & 0.7 & 0.6 & 0.6 & 0.5 & 0.6 & 0.5 \\\hline
\multirow{2}{*}{Spherical}       & Leave-one-out & 4.5  & 2.0  & 3.1 & 1.8 & 2.4 & 1.3 & 2.1 & 1.2 & 2.5 & 1.4 \\ 
                                 & All inclusive & 1.2  & 0.9  & 0.9 & 0.8 & 0.7 & 0.6 & 0.6 & 0.5 & 0.6 & 0.5 \\\hline
\multirow{2}{*}{Corner}          & Leave-one-out & 2.7  & 1.6  & 1.7 & 1.3 & 1.4 & 1.1 & 1.2 & 1.0 & 1.4 & 1.1 \\
                                 & All inclusive & 1.2  & 0.9  & 0.9 & 0.8 & 0.7 & 0.6 & 0.6 & 0.5 & 0.6 & 0.5 \\\hline
\multirow{2}{*}{\textbf{Average}}& Leave-one-out & 8.1  & 5.5  & 5.1 & 3.2 & 2.9 & 2.0 & 2.1 & 1.7 & 2.2 & 1.5 \\
                                 & All inclusive & 2.9  & 1.8  & 1.7 & 1.5 & 1.1 & 0.9 & 0.8 & 0.7 & 0.7 & 0.6 \\
\end{tabular}
\label{tab:tips}
\vspace{-1mm}
\end{table*}

First, in order to test performance for an indenter shape not seen
during training, we used a leave-one-out procedure: for any indenter
tip, we trained a regressor (similar to the one used before) on data
from all the other tips in our set, but excluding the tip being
tested. We then checked localization performance on the tip that had
been excluded from training. The second approach aimed to test
performance over multiple indenter shapes, but assuming data from each
indenter is available for training. We thus trained one model on data
from all five indenters, and tested this all-inclusive model on
separate test data from each tip. All results are shown in
Table~\ref{tab:tips}. 

We notice that high localization accuracy generalizes to multiple
indenter geometries. If data for all indenters is available for
training, \modifiedtext{3-8-2}{even light touch can be localized with
  $<$3 mm median accuracy for 4 out of 5 indenters}, further reducing to
sub-mm as contact forces increase. Even when the finger is contacted
by a never-seen-before indenter, it can localize touch well, as long
as the model has seen indenter shape variations in
training. Performance increases with normal contact force, with 2 mm
or lower median localization error typical for never-seen-before
indenters once force exceeds 0.5 N.

To illustrate the performance of contact localization and force
detection, as well as robustness for an indenter geometry not seen in
training, the accompanying video shows in real-time the predictions
made as our robotic finger is indented by an experimenter using their
own finger.

\subsection{Detection of multiple touch points}

Finally, all of our results so far assume a single touch point on the
finger. Such an assumption is applicable in some real-life cases
(e.g., a hand manipulating objects that are convex or locally convex),
but can be limiting. What happens if the finger is contacted in
multiple locations? 

Using the procedure outline in Sec.~\ref{sec:data_multi}, we recorded
a total of 1,987 data points, which we split into a 1,635-point
training corpus and a 352-point testing corpus. With this dataset, we
trained a classifier to predict, based on the tactile signals, if each
of the 20 cells was being touched. Thus, for each cell, this predictor
output a continuous probability that the respective cell was being
touched. We then tested the performance on our testing corpus. For
each entry in the test set, we considered the classification as
correct if, for every cell marked as touched or untouched in the
ground truth data, the predicted touch probability was above or below
0.5, respectively. Over our complete testing set (comprising cases
with both one and two simultaneous touches), the classification
accuracy using this rule was 96\%. When considering only cases where
the finger was simultaneously touched in two locations, the
classification accuracy was 93\%. Fig.~\ref{fig:multitouch}
illustrates the data collection procedure for this case, and shows a
number of representative examples for our touch predictor.

\section{Discussion and Conclusions}

Overall, the results confirm our main hypotheses: the overlapping
optical signal set contains the information needed to determine both
the location and normal force of the indentation with high accuracy,
and throughout the multicurved functional surface of the finger. We
can generally determine contact location with sub-millimeter accuracy,
and contact force to within 10\% (and often with 5\%) of the true
value. These results cover the complete hemispherical tip of our
finger, as well as half the circumference of its cylindrical
base. To the best of our knowledge, this level of accuracy, obtained
over the multicurved non-developable surface of a finger-shaped
package, has not been previously demonstrated.

Furthermore, these results exhibit low sensitivity to the shape of the
indenter used for contact. By training with multiple shapes we can
achieve predictors that perform well on never-before-seen indenters,
keeping the localization error below 2 mm (and in many cases below 1.5
mm) for shapes not seen during training.

Finally, our method lends itself to packaging in a form suitable for
integration into complete systems. This is due to both the overlapping
signals approach (providing a very rich signal set, with cardinality
quadratic in the number of individual sensing terminals) and our use
of time-multiplexing (further reducing the wiring needs, while
maintaining an operational frequency of 60 Hz). The results above were
obtained with a self-contained finger with a compact wiring interface
(single 14-wire FFC). Furthermore, our approach is characterized by
simple manufacturing, and accessible cost (approximately \$350 per
finger in low quantities).

\begin{figure}[t]
  \centering
    \includegraphics[clip, trim=0cm 3cm 0cm 5.8cm, width=0.5\linewidth]{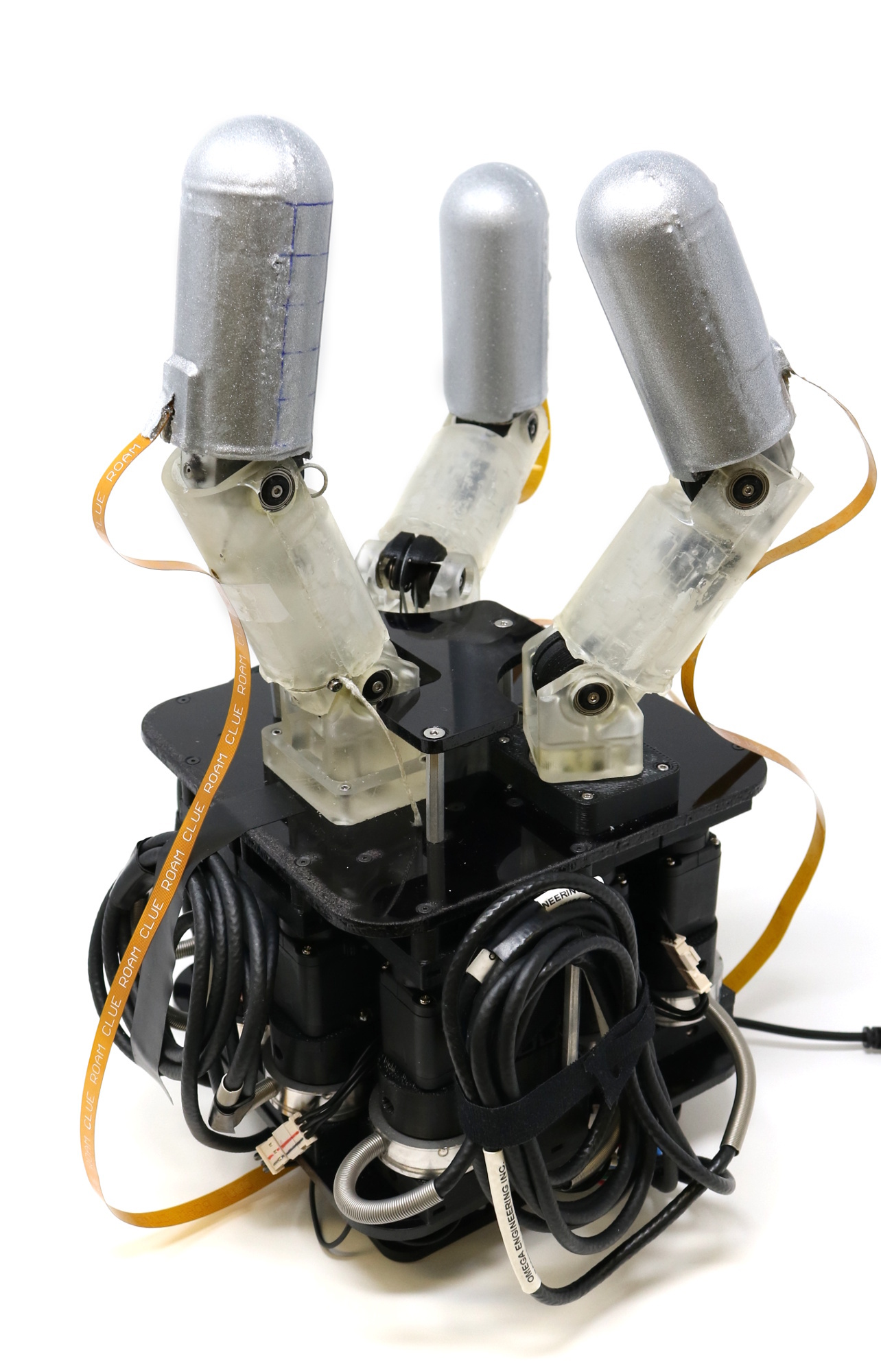}
  \caption{A robot hand with three of the tactile fingers presented
    here. The palm also houses eight motors, each equipped with a
    torque sensor.}
\label{fig:hand}
\end{figure}

\modifiedtext{3-2}{To illustrate the possibility of integration into
  a complete robot hand, in Figure~\ref{fig:hand} shows a dexterous
  robot hand using three fully integrated tactile fingers. Each finger
  is connected to a control board housed in the palm, which reads all
  tactile signals from its respective finger, as well as signals from
  motor encoders and torque sensors. This allows per-finger control
  based on both tactile and proprioceptive data running directly on
  the local control board; centralized control for all fingers must
  run on the computer that combines information from all
  three boards. However, even in this case, initial processing of the
  tactile data (e.g., going from 960 signals to contact location and
  force) can take place on the local control board.}

\subsection{Limitations}

The performance level described above results from using a purely
data-driven approach in the form of powerful regression and
classification models, which allow us to extract information from a
rich signal set without requiring analytical models. However, this
approach also comes with inherent limitations. In particular, in order
to explicitly predict a given characteristic of touch (such as
location or a force component), one must explicitly train a predictor
for it, using ground truth data. In the study presented here, this
implies the use of an instrumented setup capable of measuring ground
truth, and enough time dedicated to the collection of training data.

There are other characteristics of touch that we do not attempt to
predict in this study. This includes shear forces due to tangential
and torsional friction. Given that, unlike pressure signals, our
optical signals sense force indirectly, only through deformation of
the light-transporting medium, it is still unclear how sensitive they
will be to shear. While other tactile sensors proposed in the
literature do provide shear data, many of the most commonly used today
do not (e.g., capacitive pressure arrays). Given touch location and
normal force (both which our finger provides accurately), we believe
that a complete system can partly compensate for the lack of contact
shear force data by using torque sensing on the joints of the finger.

The most general way to encode information about a contact patch of
arbitrary shape is as a pressure distribution over the entire finger
surface. Such a method can generalize to an arbitrary number of
simultaneous contacts. \remindtext{2-5}{However, due to the difficulty of
  collecting ground truth data for such a predictor, we do not show
  this ability here.} We believe however that the information
characterizing a complete contact patch is contained in the rich
overlapping optical signal set. This hypothesis is also supported
(though not yet fully confirmed) by the ability shown in this study to
distinguish multiple touch locations. Future work must thus focus on
methods to extract and make use of this information, as we discuss
next.

\subsection{Implications for future research}

When considering future research directions, we start from the premise
that a robot tactile finger must ultimately serve as an enabler for
robotic manipulation. How does the work presented here fit in this
bigger picture?

We believe that are two main avenues for integration of such a tactile
finger into a complete manipulation system. First, model-based control
and learning methods for manipulation such as Model Predictive
Control~\cite{todorov2005,erez2012} make direct use of explicitly
derived contact properties, such as contact location and force. Under
certain conditions (e.g., locally convex objects), we have shown that
we can train machine learning models to provide such information with
very high accuracy, and over a large functional surface with no blind
spots. This can enable a wide range of methods that need this
information, e.g., to formulate and solve stability analysis or motion
equations for the hand-object system.

The second avenue, one that has recently shown tremendous promise in
complex robotic motor control problems, is that of end-to-end,
model-free sensorimotor
learning~\cite{levine2015,levine2016,levine2016b,abbeel2017,marcin2018}. These
methods learn a direct mapping from raw sensor data to motor
commands. Any intermediate representations of the sensor data are
learned at the same time as the task itself, without ground truth
other than the general reward signal related to task success. In this
context, one would use our sensors to directly train manipulation
skills, without the need to collect labeled ground truth for
intermediate representations such as contact shape or location. The
classifiers and regressors we quantify here serve to illustrate the
fact that the raw sensor data indeed comprises rich information
pertaining to contact, and suggest the sensor can be used for
end-to-end learning.

We plan to explore both of these directions in future work, using the
robotic fingers described here. \addedtext{3-2}{Possible application
  domains include grasping in extreme clutter (e.g., bin picking or
  kitting), or in-hand re-orientation of arbitrary objects for
  assembly tasks.} We believe that rich tactile information, in a
highly functional form with limited blind spots and a simple
integration path into complete systems, will prove to be an important
enabler for data-driven complex robotic motor skills, such as
dexterous manipulation.

\appendix

\noindent Finger surface parameterization: we use an
extension of a projection developed by Ro{\c{s}}ca\cite{rocsca2010},
which maps a uniform square grid in a two-dimensional $(A, B)$ space
to a near-uniform grid on a half-sphere embedded in three-dimensional
space $(x, y, z)$. This parameterization also aims to preserve local
areas as much as possible. Note that this $(A, B)$ space is
dimensionless. We use this method directly for the hemispherical
section of our finger surface (depicted in red in
Fig.~\ref{fig:geometry}), which is mapped to a square in $(A, B)$
space. The equator of the sphere thus corresponds to the outer
perimeter of the square, and each side of the square gets mapped to a
90$^{\circ}$ arc along the equator. Formally, a point in the red
region of $(A, B)$ space gets mapped to a Cartesian hemisphere of
radius $r$ centered at the origin as follows. If $0<|B|\leq|A|\leq L$:
\begin{eqnarray}
  x= \alpha \cos\left(\frac{B\pi}{4A}\right)&&
  y = \alpha \sin\left(\frac{B\pi}{4A}\right) \label{eqn:eq1}\\
  \alpha = \frac{2A}{\pi}\sqrt{\pi-\frac{A^2}{r^2}}&&
  z= r-\frac{2A^2}{\pi r}-d  
%x &=& \frac{2A}{\pi}\sqrt{\pi-\frac{A^2}{r^2}} cos\left(\frac{B\pi}{4A}\right) \label{eqn:eq1}\\
%y &=& \frac{2A}{\pi}\sqrt{\pi-\frac{A^2}{r^2}} sin\left(\frac{B\pi}{4A}\right)\\
%z &=& r-\frac{2A^2}{\pi r}-d 
\end{eqnarray}
\noindent If $0<|A|\leq|B|\leq L:$
\begin{eqnarray}
  x = \beta \sin\left(\frac{A\pi}{4B}\right)&&
  y = \beta \cos\left(\frac{A\pi}{4B}\right)\\
  \beta =  \frac{2B}{\pi}\sqrt{\pi-\frac{B^2}{r^2}}&&
  z = r-\frac{2B^2}{\pi r}-d \label{eqn:eq2}
%x &=& \frac{2B}{\pi}\sqrt{\pi-\frac{B^2}{r^2}} \sin\left(\frac{A\pi}{4B}\right)\\
%y &=& \frac{2B}{\pi}\sqrt{\pi-\frac{B^2}{r^2}} \cos\left(\frac{A\pi}{4B}\right)\\
%z &=& r-\frac{2B^2}{\pi r}-d \label{eqn:eq2}
\end{eqnarray}
where $L=r\sqrt{\pi/2}$ and $d$ is a variable we will use to extend
the mapping beyond the tip; for all points on the tip, $d=0$.

We now extend the mapping beyond the hemispherical tip and onto the
cylindrical part of our finger. Starting from the $(A,B)$ square that
maps our sphere, we extend the two sides corresponding to
180$^{\circ}$ of the equator to represent the sensorized area of the
cylindrical body. In keeping with the original projection, these new
regions are extended such that their areas in $(A, B)$ match the
frontal area of the cylinder. Formally, points in the blue or green
regions of $(A, B)$ space (Fig.~\ref{fig:geometry}) are projected into
Cartesian $(x,y,z)$ space as follows:
\begin{equation}
\label{eqn:eq3}
(A^p,B^p) = (L+A+B, -L) \quad \textrm{if $(A,B) \in $ green region}
\end{equation}
\begin{equation}
\label{eqn:eq4}
(A^p,B^p) = (L, A+B-L) \quad \textrm{if $(A,B) \in $ blue region}
\end{equation}
\begin{equation}
\label{eqn:eq5}
d = \gamma \sqrt{(A^p-A)^2 + (B^p-B)^2} 
\end{equation}
where the scaling factor $\gamma$ equals the ratio between the real
cylinder height and L2 norm of the segment where the green region
meets the blue region. The projected values $(A^p,B^p)$ and $d$ are
plugged back into Eqs.~(\ref{eqn:eq1}-\ref{eqn:eq2}) to obtain the
Cartesian $(x,y,z)$ values for points on the cylindrical area of
the finger.

\section*{Acknowledgment}

This work was supported in part by the National Science Foundation under grants CAREER IIS-1551631 and NRI CMMI-1734557.

\ifCLASSOPTIONcaptionsoff
  \newpage
\fi

\bibliographystyle{IEEEtran}
\bibliography{bib/tactile_sensing}

\vspace{-10mm}
\begin{IEEEbiography}[{\includegraphics[width=1in,height=1.25in,clip,keepaspectratio]{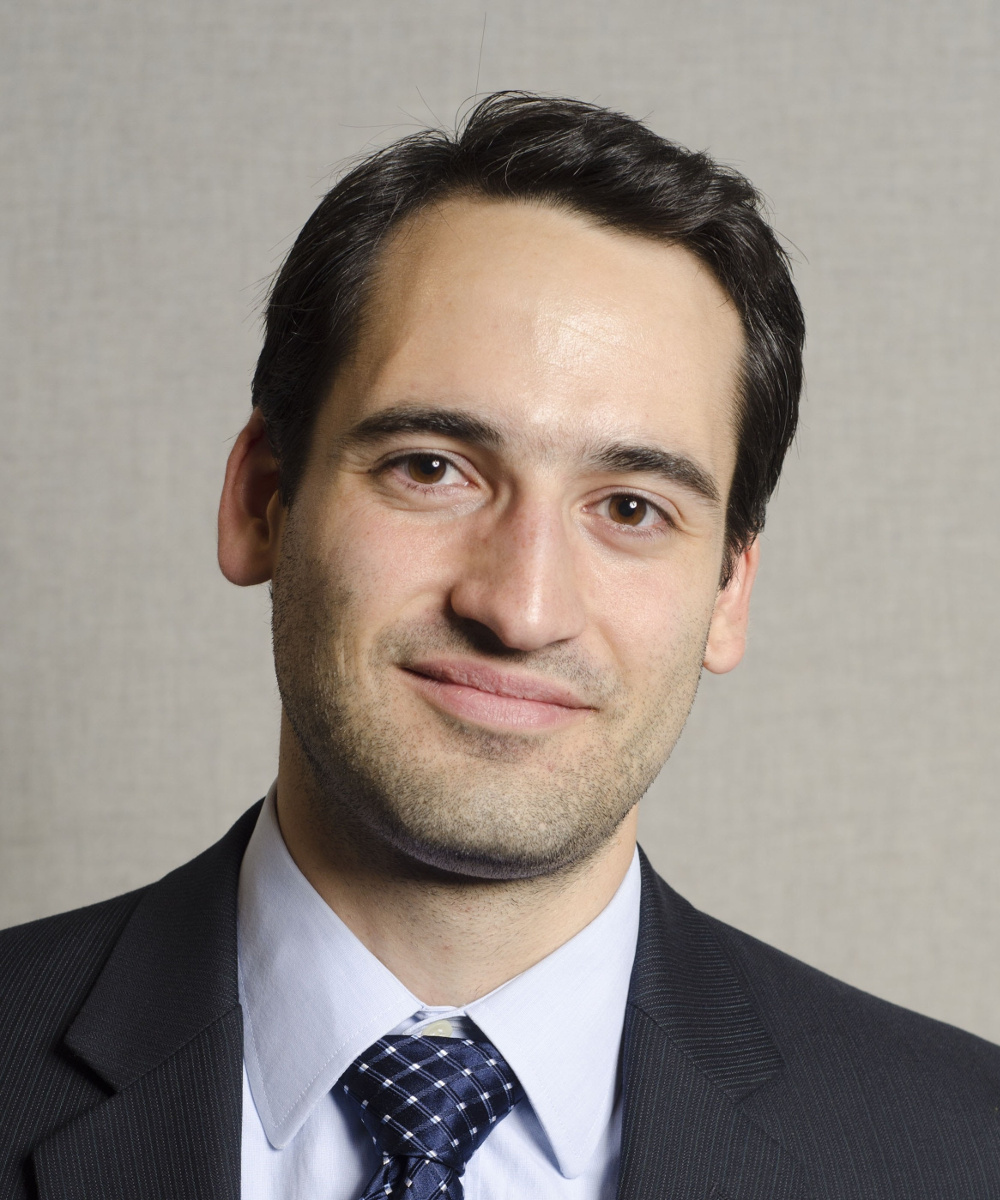}}]{Pedro Piacenza}
  is a Ph.D. candidate at Columbia University. He obtained his M.S. from
  Columbia University, and his B.S. in Electrical Engineering from
  Universidad Catolica Argentina. His research focuses on developing
  tactile feedback systems for robotic manipulation.
\end{IEEEbiography}
\vspace{-15mm}
\begin{IEEEbiography}[{\includegraphics[width=1in,height=1.25in,clip,keepaspectratio]{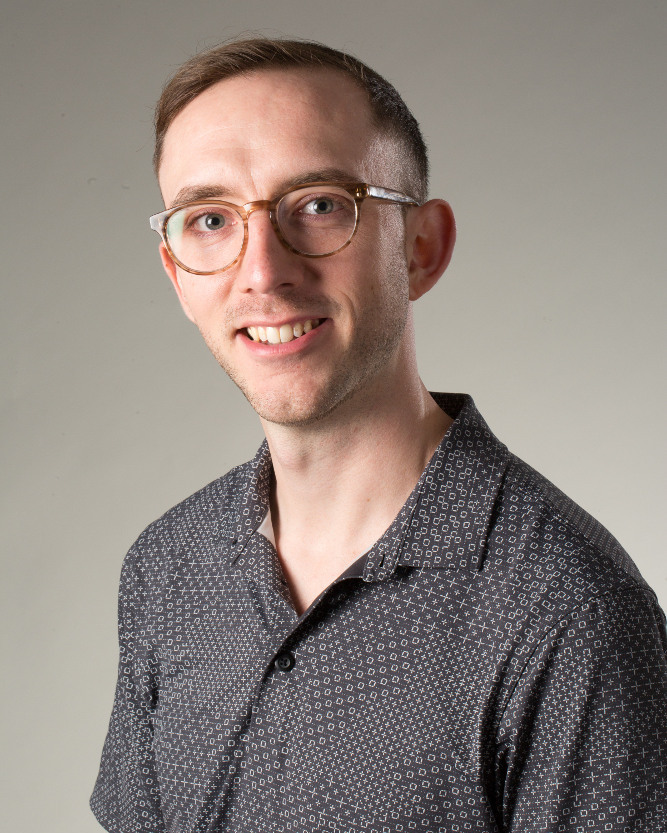}}]{Keith Behrman}
   is a Ph.D. candidate in the Columbia Lab for Unconventional
   Electronics (CLUE) at Columbia University. He obtained his M.S from
   Columbia and his B.S. in Physics from The University of
   Minnesota. His research is focused in micro-LED devices for
   next-generation display systems.
\end{IEEEbiography}
\vspace{-15mm}
\begin{IEEEbiography}[{\includegraphics[width=1in,height=1.25in,clip,keepaspectratio]{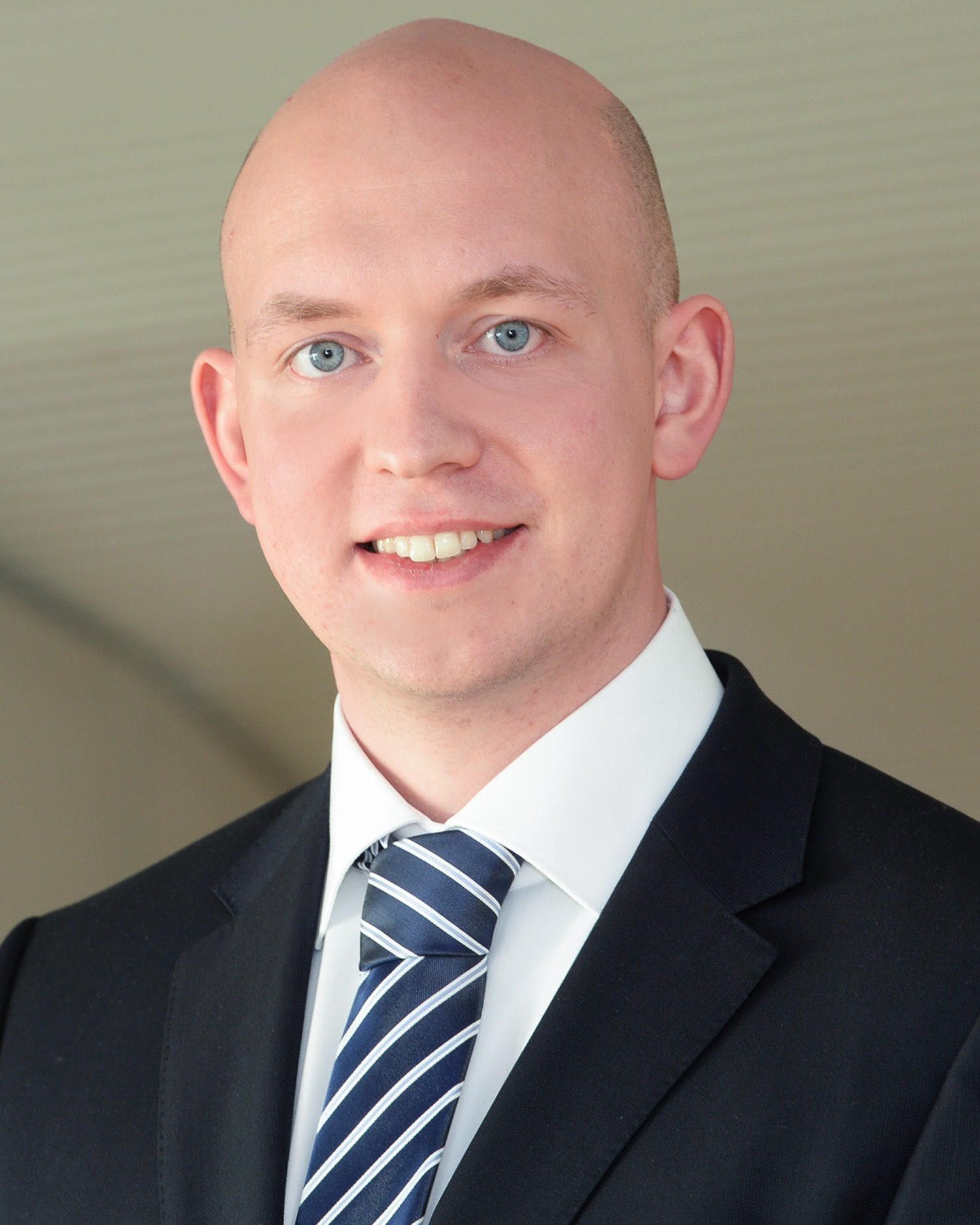}}]{Benedikt Schifferer}
  is an M.S. candidate in Data Science at Columbia University. He
  obtained his BSc. in Mathematics and Business Administration at the
  University of Mannheim.
\end{IEEEbiography}
\vspace{-15mm}
\begin{IEEEbiography}[{\includegraphics[width=1in,height=1.25in,clip,keepaspectratio]{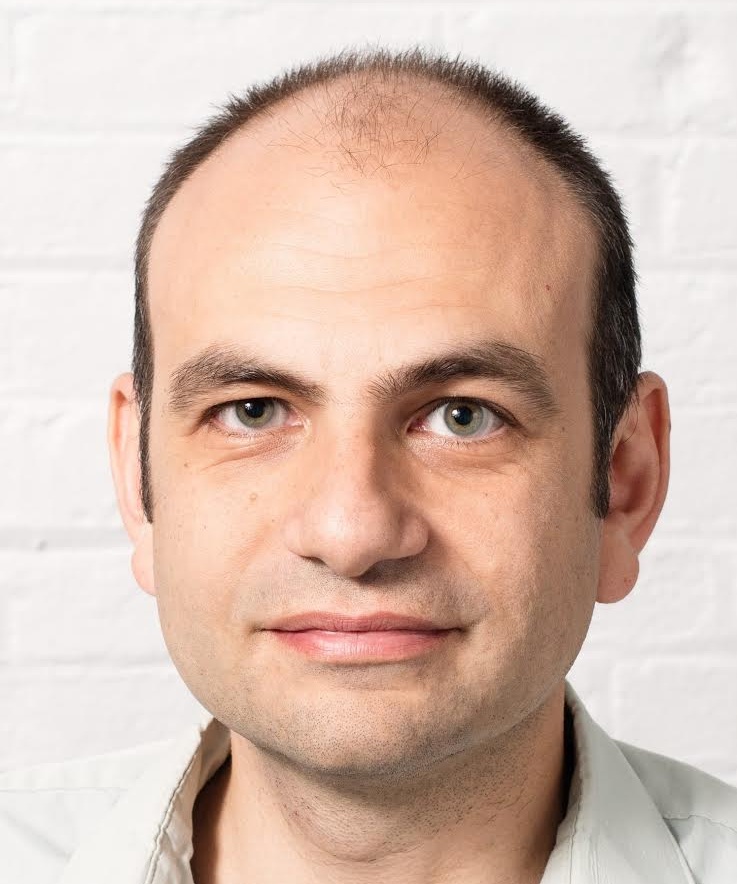}}]{Ioannis Kymissis}
is Professor in the Electrical Eng. Dept. at Columbia University. He
obtained his S.B., M.Eng. and Ph.D. degrees from MIT, and also
participated in a cooperative program through which he completed his
M.Eng. thesis at IBM Research. His research focuses on fabrication and
characterization of thin film electronics.
\end{IEEEbiography}
\vspace{-15mm}
\begin{IEEEbiography}[{\includegraphics[width=1in,height=1.25in,clip,keepaspectratio]{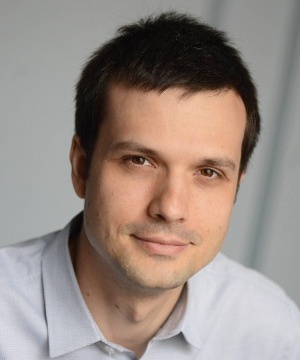}}]{Matei Ciocarlie}
  is Associate Professor at Columbia University. He completed his
  Ph.D. and M.S. degrees at Columbia University, and his B.S. at the
  Polytechnic University of Bucharest. His main research interest is
  in reliable robotic performance in unstructured, human environments,
  looking to discover how artificial mechanisms can interact with the
  world as skillfully as biological organisms.
\end{IEEEbiography}

\end{document}